\documentclass{article}

\usepackage[preprint]{corl_2021} 

\usepackage{times}

\usepackage[numbers]{natbib}
\usepackage{multicol}

\usepackage{amsmath} 
\usepackage{amssymb}  
\usepackage{breqn}
\usepackage{bm}

\usepackage[utf8]{inputenc}
\usepackage{booktabs}
\usepackage{multirow}
\usepackage{subcaption} 
\usepackage[skip=0pt,font=small,labelfont=bf]{caption}
\usepackage{array}
\usepackage{xspace}
\usepackage{algorithm}
\usepackage[noend]{algpseudocode}
\usepackage{wrapfig}

\hypersetup{urlcolor=blue, colorlinks=true}
\usepackage[colorinlistoftodos,prependcaption,textsize=tiny]{todonotes}

\definecolor{international_orange}{RGB}{240, 74, 0}
\newcommand{\revised}[1]{#1}
\newcommand{\eat}[1]{}
\title{Predicting Stable Configurations for \\Semantic Placement of \revised{Novel} Objects}

\author{Chris Paxton\(^{1}\), Chris Xie\(^{2}\), Tucker
   Hermans\(^{1,3}\), Dieter Fox\(^{1,2}\)
   \thanks{$^{1}$NVIDIA Corporation, Seattle, USA;$^{2}$Paul G. Allen
     School of Computer Science \& Engineering, University of Washington,
     Seattle, USA;$^{3}$School of Computing, University of Utah, Salt
     Lake City, USA}%
}
\begin{document}

\maketitle

\begin{abstract}
Human environments contain numerous objects configured in a variety of arrangements. Our goal is to enable robots to repose previously unseen objects according to learned semantic relationships in novel environments. We break this problem down into two parts: (1) finding physically valid locations for the objects and (2) determining if those poses satisfy learned, high-level semantic relationships.
We build our models and training from the ground up to be tightly integrated with our proposed planning algorithm for semantic placement of unknown objects. We train our models purely in simulation, with no fine-tuning needed for use in the real world.
Our approach enables motion planning for semantic rearrangement of unknown objects in scenes with varying geometry from only RGB-D sensing.
\revised{Our experiments through a set of simulated ablations demonstrate that using a relational classifier alone is not sufficient for reliable planning. We further demonstrate the ability of our planner to generate and execute diverse manipulation plans through a set of real-world experiments with a variety of objects.}

\keywords{Deep learning for manipulation, learning for motion planning, semantic manipulation}


\end{abstract}

\begin{figure}[h]
\vskip -12pt
\centering
\includegraphics[width=\textwidth]{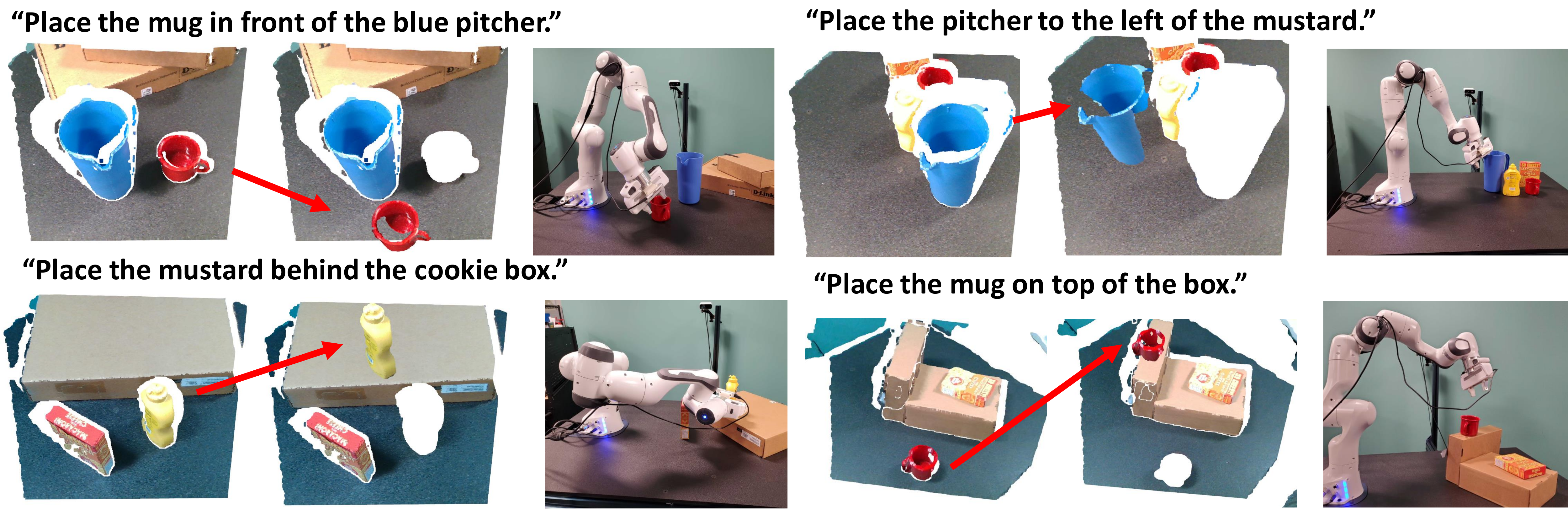}
\vskip 6pt
\caption{Planning to achieve different predicate relationships in the real world, given only segmented point clouds of the scene. The planner computes final placement positions and associated kinematic poses, which lets the robot generate a motion plan to place these objects at different locations in the real world.}
\label{fig:cover-pairs}\vspace{-12pt}
\end{figure}

\section{Introduction \& Motivation}\label{sec:intro}


Many robotics tasks in human environments involve reasoning about the relationships between different objects and their locations in a particular environment.
Imagine a robot tasked with pouring a cup of coffee, it must reason about the
relative position of the cup to the coffee pot, such that it pours the
coffee into the cup and not onto the counter or floor. For a robot
attempting to brew the coffee prior to pouring, many more multi-object
relations must be reasoned about for the tasks of retrieving the
necessary equipment from storage, scooping and pouring beans or
grounds, and filling any required water vessels. 

Classically, researchers investigating these sorts of multi-object
planning and manipulation problems assume full knowledge of the
objects and their poses in the scene~\cite{garrett2020integrated}.
However, for robots acting in homes and other
open-world environments, it is unreasonable to assume that a deployed
robot will have knowledge and accurate pose estimation of all objects
in the environment. Furthermore, simple hand-engineered classifiers for determining logical predicates often exhibit unintended behavior when deployed on real-world systems~\cite{Srinivasa-2016-5607}. For these reasons, rearrangement has recently been identified as a major challenge for robotics~\cite{batra2020rearrangement}.

Given these challenges, its natural to examine the use of learning to improve task and motion planning with real sensing~\cite{xu2019regression,nerp-rss2021,huang2019continuous, paxton2019visual, kase2020transferable,zhu-icra2021-graph-planning}.
However, previous methods fail to solve the full problem of unknown object rearrangement with physical robots. Some only operate on known objects~\cite{huang2019continuous,zhu-icra2021-graph-planning}, others ignore or significantly restrict the space of robot control~\cite{xu2019regression, nerp-rss2021} or relations~\cite{paxton2019visual,kase2020transferable}, while still others make assume an explicit goal configuration is given~\cite{nerp-rss2021}. An alternative approach to solve complex manipulation tasks relies on learning model-free neural net policies instead of explicit models of conditions and effects~\cite{huang-cvpr2019-ntg, xu2020deep}. Such methods have so far failed to show the level of generalization across objects and environments capable by modern task and motion planners.

\revised{In this paper, we focus on a critical subtask of rearrangement planning--\emph{semantic placement}--where the robot must perform a pick and place operation to move an object into a stable configuration that satisfies a desired set of semantic relations. This allows us to focus on three specific objectives not addressed in previous work on learning for task and motion planning.} First, we examine the problem of planning and controlling manipulation behaviors to change inter-object relations of potentially previously unseen objects using RGB-D sensing. Second, we aim to learn the necessary relations from data to avoid the bias introduced from hand-engineered classifiers. Finally, we must learn what realistic scenes look like--so that the robot can ensure that it places objects in reasonable locations which are stable and free of undesired collisions.

Our work relies on the ability to infer inter-object
relational predicates between objects proposed from an RGB-D instance
segmentation, e.g.~\cite{xiang2020learning}.
Semantic relations serve an important role in instructing robots~\cite{forbes2015robot}. As such, researchers have examined visual prediction of spatial relations~\cite{rosman-ijrr2011-relations,clement-pr2018-spatial-relations,mees2020learning,bear2020learning} including inference of support relations~\cite{panda-iros2013-support-order,zhang-iros2019-support-relations}.
While these methods can be sophisticated incorporating language~\cite{paul2016efficient,venkatesh2020spatial,mees2020learning} or scene graph information~\cite{sharma2020relational,wilson-corl2019-collection-pushing,bear2020learning}; none have demonstrated the ability to integrate prediction with robot planning and control to satisfy new relations between unknown objects. \revised{Indeed, we show semantic prediction alone is insufficient for reliably planning successful semantic rearrangement.}

\revised{
In addition to this semantic prediction, we need to identify \textit{physically realistic} poses where these objects can be placed.
Stable placement prediction has also received attention from robotics researchers~\cite{jiang-ijrr2012-placement,Place-Henrich-GeometricGpuBased2014,Place-StoneStack2017,Place-DrummondRotationStabilitylearning2021}; however, learning-based approaches~\cite{jiang-ijrr2012-placement,Place-DrummondRotationStabilitylearning2021} train directly to maximize stability of object placement.
In contrast our approach simply distinguishes realistic configurations from invalid ones, 
which allows us to learn a general-purpose \textit{scene-realism discriminator} which can capture wide distributions over realistic poses in 3D space.
}
This provides the benefit of a more general model for use in arrangement planning, while giving up somewhat the level of precision seen in some placement-specific methods~\cite{jiang-ijrr2012-placement, Place-StoneStack2017}.
\revised{In addition, it means we can use
simulation-based data to train both our relational prediction and scene discriminator directly on raw point cloud and associated segmentation masks, and transfer to the real world.}

We embedded our learned relational predictor and scene discriminator within a
sampling-based planning framework to change relations in the
scene in a goal directed way. By planning directly over changes in
segment pose in a point cloud, we can decouple the goal generation
problem from the robot control problem, allowing us to leverage
model-based, state-of-the-art motion planning and grasp
prediction algorithms~\cite{lu2020multi, sundermeyer2021contactgraspnet} to perform
the necessary manipulation. We further accelerate our planner by learning an object pose sampler conditioned on desired relations to initialize the optimization, similar to the grasp planning approach in~\cite{lu2020multi}.
Production of these goal states can then be used for semantically-defined
placement tasks.

\revised{We highlight the advantages of our approach over a variety of baselines and ablations of our full method. Critically, we demonstrate that both the relational classifier and scene discriminator are necessary for reliably generating successful plans.} 
We then demonstrate our approach in the real world using a
Franka robot (Fig.~\ref{fig:cover-pairs}). Our experiments constitute the first physical-robot demonstration that combine learned models for inter-object relations and stability estimation enabling rearrangement of novel objects.
\revised{Crucially, we show that the combination of relationship classifier and scene discriminator allows us to plan placements for a variety of relationships in cluttered scenes.}
In addition, our models are trained entirely in simulation with no need for real-world fine tuning.





\section{Methods}\label{sec:methods}
Given a single view of a scene containing objects for which our robot potentially has no previous experience, we wish for the robot to rearrange the scene to satisfy some new set of logical constraints. For example the robot may be tasked to move the \emph{query object} \(i\) to be on the far right side of \emph{anchor object} \(j\) or to be stacked on top of object \(k\). Each individual relationship between $i$ and $j$ is referred to as a predicate $\rho_{ij}$; we can describe multiple logical relationships as the vector $\vec{\rho}_{ij}$.

We assume we are given a partial-view point cloud \(Z\) with segment labels for each point to identify the different objects. Given this point cloud and a set of logical predicates describing the desired relationships $\vec{\rho}_{ij}$, the robot must find a pose offset $\delta$ (3D translation and planar rotation) for object $i$ that satisfies $\vec{\rho}_{ij}$ and is additionally a stable, physically valid placement pose in the environment.

Thus there are two key components in our rearrangement and placement planning approach: predicting which poses objects can be physically placed and predicting which poses satisfy the given high-level instructions. We formalize predicate planning as the following problem:
\begin{flalign}
\underset{\delta}{\text{argmin}}\hspace{15pt}&     c(\delta) = \lambda_f \|1 - f(x_i \oplus \delta, Z')\|_2 + \lambda_\rho \| p_\rho(x_i \oplus \delta, x_j) - \vec{\rho}_{ij}\|_2 \label{eq:planning_problem}\\
\text{subject to} \hspace{8pt} & T(Z, x_i \oplus \delta) = Z' \label{eq:transform-constraint} \\
  & f(x_i \oplus \delta, Z') > \epsilon_f \label{eq:real-constraint}\\
  &p_\rho(x_i \oplus \delta, x_j)\left[\vec{\rho}_{ij}\right] > \epsilon_\rho \label{eq:predicate-constraint} \\
  &\Pi(x_i \oplus \delta) = 1 \label{eq:in-view-constraint}
\end{flalign}
where $x_i$ and $x_j$ are the object point clouds for objects $i$ and $j$, respectively, and $\oplus$ denotes the application of the 3D translation and planar rotation.
At the heart of our planning cost, Eq.~\ref{eq:planning_problem}, are two models. The first $f(x_i \oplus \delta, Z')$ is a neural net trained to determine if the resulting scene is physically realistic and stable. 
The second term defines the cost associated with matching the set of target predicates, where $p_\rho(x_i \oplus \delta, x_j)$ estimates the set of predicate relationships between the transformed point cloud $x_i \oplus \delta$ and $x_j$. This cost implies maximizing the likelihood that the resulting scene is both realistic and satisfies the desired predicates. 

In addition to the cost function, we put minimum bound constraints on the physical feasibility (Eq.~\ref{eq:real-constraint}) and predicate probabilities (Eq.~\ref{eq:predicate-constraint}). We use $\vec{\rho}_{ij}$ as an index in Eq.~\ref{eq:predicate-constraint} to extract the subset of predicted relations that must be satisfied.
This ensures we never attempt to plan to a scene configuration with low probability of success, even if it defines a local optimum of the objective.
Eq.~\ref{eq:transform-constraint} models the transition of applying offset $\delta$ to $x_i$ in the observed scene $Z$ to generate the resulting scene \(Z'\), which we evaluate in the cost and other constraints. Finally Eq.~\ref{eq:in-view-constraint} ensures that sampled object offsets are visible in the camera view of the robot, since our cost and constraint evaluations would be ill-defined otherwise. We describe the details of these models and their construction in Sec.~\ref{sec:models} and give a detailed description of our data generation process for training in Section~\ref{sec:dataset}.

We use a variant of the cross-entropy method (CEM)~\cite{kobilarov2012cross} to solve this constrained optimization problem. CEM has previously been applied to robot motion planning~\cite{kobilarov2012cross}, including semantic motion planning from learned models~\cite{paxton2016want}. We provide further details of our planner in Section~\ref{sec:planner}.


\begin{figure*}[bt]
\includegraphics[width=\textwidth]{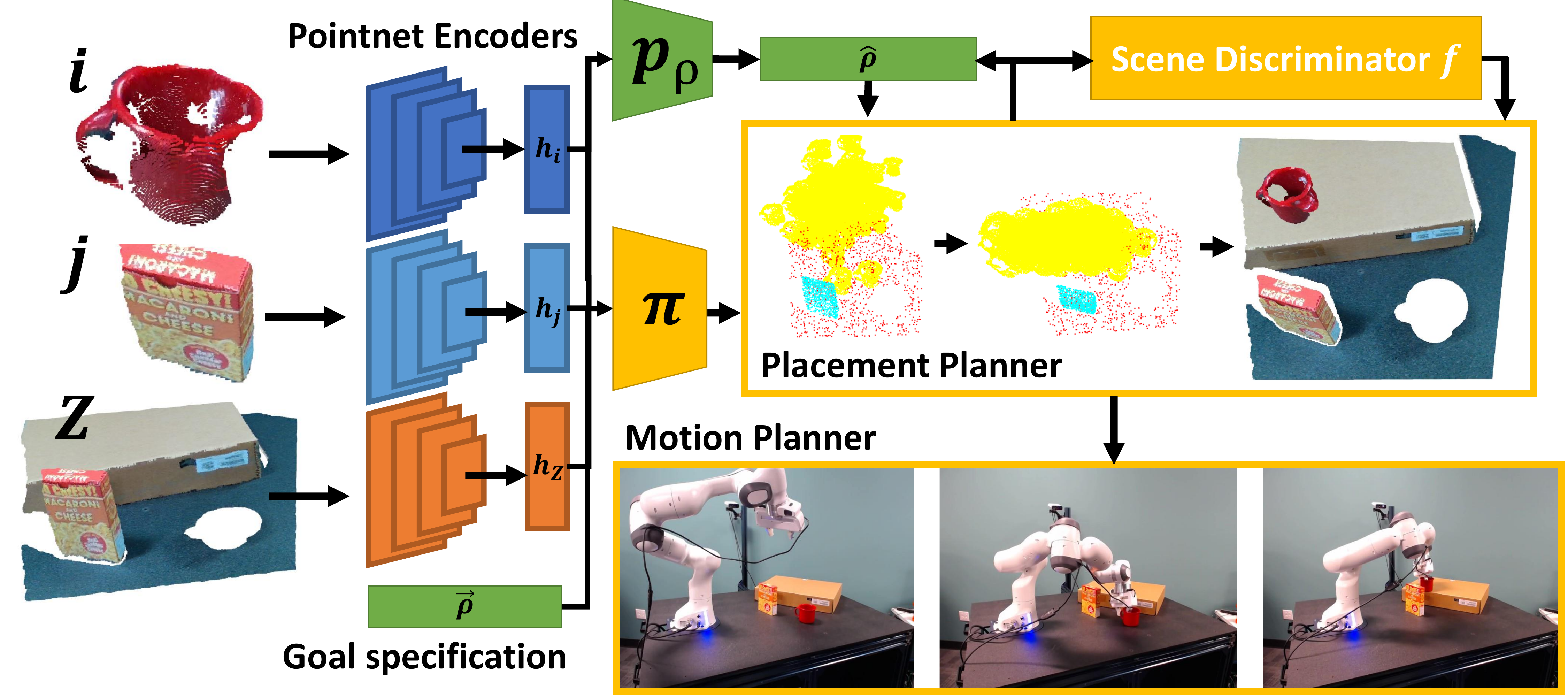}
\vskip 8pt
\caption{System for rearrangement of unknown objects according to spatial relations, where we wish to move a query object $i$ (in this case a red mug). We encode objects with Pointnet++ to predict predicates, based on the region around an ``anchoring'' object, $j$ (the macaroni box) to which the query is relative. Query locations are sampled based on a learned prior $\pi$, and are classified by the scene discriminator network $f$ as either realistic or unrealistic. This is used as a part of an optimization algorithm to find a stable, kinematically feasible query pose so that we can place the query object $i$ in a new location.}
\label{fig:system}
\vskip -12pt
\end{figure*}

\subsection{Rearrangement Planner Models and Training}
\label{sec:models}
We train multiple neural networks to compute the values needed to instantiate our planning problem defined by Equation~\ref{eq:planning_problem}: the current set of predicates,
and the discriminator score which describes whether or not a particular set of object points $x$ defines a realistic configuration in the scene. 

\textbf{Object and scene encoders} The core piece of the model is a PointNet++~\cite{qi2017pointnet++} encoder which extracts a lower-dimensional object representation $h$. Each network takes two objects $i$ and $j$ represented as point clouds $x_i$ and $x_j$, as well as the scene point cloud $Z$, as input. Given an observation point cloud $x$, we learn a mapping $e(x) \rightarrow h$ for the objects, in order to get two latent representations $h_i$ and $h_j$ for the query and anchor objects.
The object encoder is a Pointnet++ model with three set abstraction layers; see the supplemental material for further details.

We train a separate scene encoder $e_Z(x_Z)$ \revised{to capture the objects' relation to other scene geometry when predicting where it can be placed. This outputs} $h_Z$ encoding scene-specific information, where $x_Z$ is centered around the anchor point $o_j$ -- the centroid of $x_j$. Points are sampled in a radius of $r=0.5$m around $o_j$.
We choose this representation to define features from the fixed perspective of the anchoring object $j$.

\textbf{Relation predictor} The predicate classifier network $p_\rho(x_i, x_j) \rightarrow \hat{\rho}_{ij}$ estimates which predicates are true for a pair of object point clouds.
It uses the representations from the object encoder, $e(x_i) = h_i$ and $e(x_j) = h_j$, for the query and anchor objects $i$ and $j$, respectively.
These are passed into an MLP which predicts a vector of length $N_{predicates}$. When used as part of the planning algorithm, we ignore predicted predicates that are not part of the goal specification.

\textbf{Pose prior} In addition, we learn a prior distribution over possible poses where the object might satisfy the predicates, relative to the anchoring object. 
We implement this prior $\pi(x_i, x_j, Z, \vec{\rho}_{ij}) \rightarrow \{\alpha_k, \mu_k, \sigma_k\}_{k=1:K}$ as a Mixture Density Network (MDN)~\cite{bishop1994mixture}, which predicts the parameters of a Gaussian mixture model with \(K\) components. This GMM distribution defines the probability of a specific pose offsets $\delta$ with respect to the anchor point $o_j$. The relationships are assumed to be spatially defined relative to this object, so that the predicted center of the query object $i$ is $\hat{o}_i = o_j \oplus \delta$. Once trained we can produce samples from the MDN by evaluating it for the current observations and then applying standard GMM sampling using the output parameters.

As with the relation predictor $p_\rho$, the pose prior uses the object encoder to get lower-dimensional representations $h_i$ and $h_j$ for each object as well as the scene encoder to predict $h_Z$.

\textbf{Scene discriminator} The relation predictor and prior on their own are not enough to find stable placement poses.
As such, we define a discriminator network $f(x, Z)$ trained to predict if a given configuration results in a placement that is physically realistic w.r.t. the training data, i.e. it is stable.

This model uses a slightly different architecture from the above models, and unlike them we first center the scene on $\hat{o}_i$, the query point and potential new pose for object $i$.
We look at the local region around $\hat{o}_i$ with a fixed radius $r=0.5$m to classify whether a particular $\delta$ would result in a realistic placement pose.
We use a single PointNet++ model to encode points from both objects together, given a label indicating which points belong to the query object (the object whose placement we are attempting to find).
In practice, the sphere-query necessary to extract the local context around a particular pose is the same as that used in PointNet++, so this operation can be performed quickly at inference time.

\textbf{Transform operator} We require one additional operator to plan on point clouds: the transformation operator $T(Z, x_i \oplus \delta)$. We assume a deterministic transition function and rigidly transform points associated with the query object according to the relative pose offset \(\delta\) constructing a new scene $Z'$.

\textbf{Model Training} We jointly train the object and scene encoders $e$ and $e_Z$, relationship predictor $p_\rho$, and the pose prior $\pi$. The relationships can be directly supervised from our training data, given knowledge of the ground truth predicates, and are trained with a binary cross entropy loss.
The prior $\pi$ is trained to predict the offset $\delta$ from the anchor point $o_j$ to the observed pose of the query object, $o_i$, in the training data; i.e. $\delta = o_j \ominus o_i$. 

To train the scene discriminator, we first note that all of our training data consists of stable object placements, thus there is no negative data nor stability supervision available. Instead, we create negative data online by applying random $\delta$ offsets to the pose of the query object $i$.
We simply sample a random $\delta$ and apply our transform operator to create a new scene $Z^- = T(Z, x_i \oplus \delta)$. These $\delta$ were sampled to be between $2$ and $15$ centimeters in a random direction. The resulting scenes are highly likely to not be physically realistic nor stable (e.g. object $i$ is floating or is in collision).

\begin{algorithm}[h]
\caption{Placement planning algorithm pseudocode.}
\label{alg:placement}
\begin{algorithmic}[1]
\Function{FindPlacement}{object point cloud $x_i$, object point cloud $x_j$, scene $Z$, goal $\vec{\rho}_{ij}$}
\For{$s \in \text{range}(1,N)$}
    \State $\vec{\delta}_s = \emptyset$
    \While {$\text{length}(\vec{\delta}_s) < B$} \Comment{Rejection Sampling} \label{alg:rejection-start}
        \If{$s = 0$}
            \State $\delta \sim \pi\left(x_i, x_j, Z, \vec{\rho}_{ij}\right)$ \Comment{Sample initial poses from learned prior}
        \Else
            \State $\delta \sim \mathcal{N}(\mu', \Sigma')$ \Comment{Sample subsequent poses from surrogate distribution}
        \EndIf
        \State $Z' \leftarrow T(Z, x_i \oplus \delta)$ \Comment{Shift object point clouds by $\delta$}
        \State $\texttt{realistic} \leftarrow f(x_i \oplus \delta, Z') > \epsilon_f$ \Comment{Determine if pose is realistic}
        \State $\texttt{goal} \leftarrow p_\rho(x_i \oplus \delta, x_j)\left[\vec{\rho}_{ij}\right] > \epsilon_\rho$ \Comment{Classify if goal predicates are true}
        \State $\texttt{in\_view} \leftarrow \Pi(x_i \oplus \delta)$ \Comment{Ensure it will be visible}
        \If{\texttt{realistic} and \texttt{goal} and \texttt{in\_view}}
        \State $\vec{\delta}_s \leftarrow \vec{\delta} \cup \delta$ \Comment{Add new $\delta$ to batch of samples}
        \EndIf
    \EndWhile
    \State Compute cost \(c(\delta)\) for each $\delta \in \vec{\delta}_s$
    \State Sort $\vec{\delta}_s$ and take top $N_{\texttt{elite}}$
    \State Fit surrogate distribution parameters $(\mu', \Sigma')$
\EndFor
\State
\Return lowest-cost $\delta$ seen so far, or $\emptyset$
\EndFunction
\end{algorithmic}
\end{algorithm}

\vspace{-18pt}
\subsection{Manipulation Planning with Relationship Models}
\label{sec:planner}
We now describe how we solve the optimization problem for Equation~\ref{eq:planning_problem} using the components defined above. Alg.~\ref{alg:placement} describes how we can place an object so as to satisfy a particular relationship.
We take as given a set of desired relationships $\vec{\rho}_{ij}$ in scene $Z$, and target objects $i$ and $j$, where $i$ is the \emph{query object} that we will be moving and $j$ is the \emph{anchor object} that will be kept stationary.

Initially, we perform rejection sampling to draw a batch of $B$ candidate pose offsets from our MDN prior \(\pi(x_i, x_j, Z, \vec{\rho}_{ij})\) (line 6) keeping only those that satisfy the full set of constraints (lines 10--14). We sample until either a time budget has been reached or $B$ samples have been successfully drawn.

\begin{wrapfigure}{r}{0.5\textwidth}
  \vspace{-12pt}
\includegraphics[width=\linewidth]{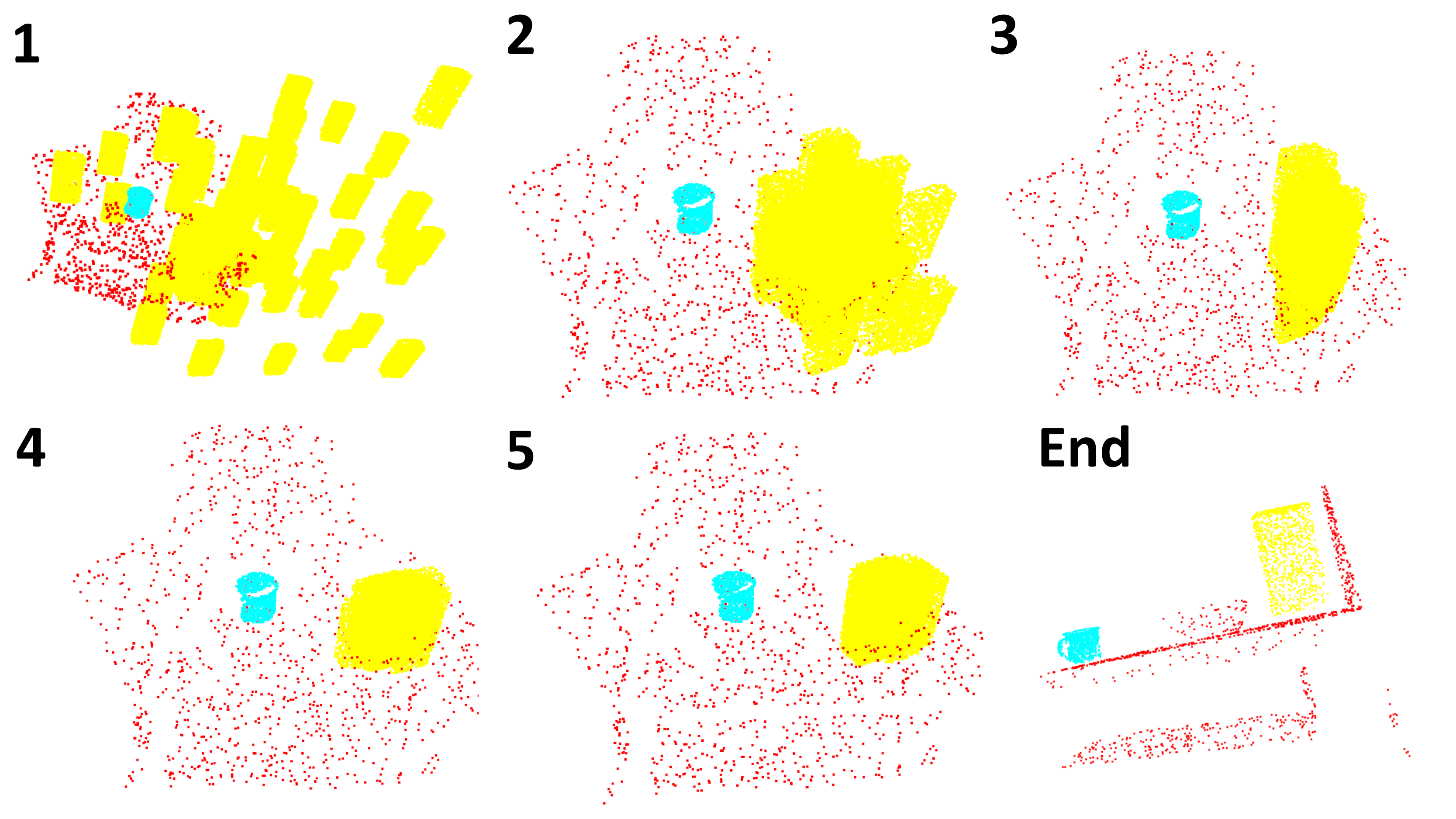}
\caption{Depiction of the planning sequence. Initial samples are drawn from an MDN prior $\delta \sim \pi(Z, \vec{\rho}_{ij})$. Final poses satisfy both realism and predicate constraints finding a reasonable placement pose.}
\label{fig:sampling}
\end{wrapfigure}
For each sample \(\delta\) we compute a new query point cloud using the transform function $Z' = T(Z, x_i \oplus \delta)$ (line 9) and thus satisfy the constraint in Eq.~\ref{eq:transform-constraint} by construction. We then evaluate the remaining constraints in Eq.~\ref{eq:real-constraint}--\ref{eq:in-view-constraint} (lines 10--12) accepting only feasible samples (lines 13--14).
Next, as per the cross-entropy method, we evaluate the cost of the valid samples (line 15) and fit a surrogate distribution to the best scoring \(N_{\texttt{elite}}\) samples with mean $\mu'$ and variance $\Sigma'$ (lines 16--17).
At each subsequent step, we draw $B$ samples $\delta \sim \mathcal{N}(\mu', \Sigma')$ from our current surrogate distribution, compute the scores again, and re-sample, until we have performed \(N\) sampling iterations.

Fig.~\ref{fig:sampling} gives an example for how the planning algorithm works in practice to achieve the goal that the yellow object be aligned to the right of the cyan object.
\revised{In both Fig.~\ref{fig:system} and Fig.~\ref{fig:sampling}, yellow objects are predicted positions of the query object; cyan represents the anchor object, and red points represent downsampled scene geometry. This is an accurate depiction of the inputs into our model. The final frame of Fig.~\ref{fig:sampling} has been rotated to show the precise alignment with the table surface.}

\subsection{Dataset Creation}
\label{sec:dataset}
We generated a large-scale dataset in simulation of RGB-D images with associated segmentation masks and relational predicates. We provide binary labels between all visible pairs of objects for all predicates listed in Table~\ref{tab:dataset-results}. 
Each scene consists of 3 to 7 random Shapenet~\cite{shapenet2015} objects in stable configurations on various surfaces, including in stacks. We include mugs, bowls, and bottles, as well as boxes and cylinders of random sizes. \revised{Examples of generated scenes are shown in the supplemental material}. 
\revised{Objects were placed in random configurations, and we ran physics forward to find stable arrangements.} We then rendered images both with and without each object. In order to train the rotation model, we render each object on its own and apply a random rotation. 
Directional predicates (\textit{left of}, \textit{right of}, etc.) are computed based on bounding box overlap; others are computed based on distance between meshes or ground-truth position and orientation. See the supplementary materials for further details.



\section{Experiments}\label{sec:experiments}

We first performed a set of simulation experiments on scenes that resembled our real world objects examining different versions of our model. 
\revised{First, we show an ablation test of our method on held-out YCB objects~\cite{calli2015ycb}, with a set of known grasps from Eppner et al.~\cite{eppner2019billion}, in order to show that our algorithm with discriminator is better able to find stable positions for objects and scenes that did not appear in our training data. Second, we show a break-down of the predicate results, showing that our learned model is comparable or better at capturing a wide range of difficult relationships even in partially-occluded scenes.}







 

We sampled 100 random scenes in the kitchen environment from Fig.~\ref{fig:sim-experiments}, with objects either positioned on top of the counter or in the top drawer. These objects did not appear in the training set.
Table~\ref{table:planner} shows a breakdown of several different versions of the planner after running experiments on 100 random scenes, each with a random predicate goal chosen from (in front, behind, left, right). Due to the random placements of the objects, not all scenes are feasible, and in many cases the goal pose would be occluded or off of the table to the front, resulting in challenging planning problems.
\begin{table*}[hbt]
    \centering
    \begin{tabular}{l c c c c c c}
    \toprule
    Variant & Found Predicates & Found Realistic & Successful & Stable Pose \\
    \midrule
    Full model, $\lambda_f = 100$ & 93	& 87 & 84 &	71 \\
    Full model, $\lambda_f = 1$ & 94 & 81 & 78 & 67 \\
    No Discriminator ($\lambda_f = 0$) & 100 & 3 & 3 & 25 \\
    Mean only & 96 & 27 & 27 & 39 \\
    MDN Prior &	99 & 6 & 6 & 42 \\
    \bottomrule
    \end{tabular}
    \vskip 6pt
    \caption{Comparison between different versions of the planner, when tested on 100 random multi-object scenes in a kitchen environment. ``Stable pose'' is when the object center moved less than 5 cm after placement.}
    \label{table:planner}
    \vskip -12pt
\end{table*}

The different baselines look at the effects of the discriminator model, which determines whether or not a scene is realistic and whether or not an object can be placed at a particular pose.  For example, the ``no discriminator'' case does not use the discriminator at all, and is very good at finding poses matching the predicate goal but not finding stable poses. We also vary the weight of the discriminator $\lambda_f$ in several examples. For these experiments we use a batch size, $B$, of $100$.


We ran two experiments that do not use the discriminator in our ``full'' planning approach.
\textbf{Mean only} draws samples only from the mean of the MDN prior $\pi$. This looks at what performance is like with a learned policy. 
\textbf{MDN Prior} uses the learned mixture density function as a cost function in place of using the discriminator, since presumably this might capture much of the same information. We can see that it actually does a fairly good job at matching the predicates, but is not very discriminative when it comes to finding stable poses for placement. Both of these perform notably worse at finding stable poses in our test environments.

\subsection{Scene Discriminator Performance}


Here, we examine the performance of our scene discriminator. 
To do this, we compare placement poses sampled from the MDN prior distribution in randomly-generated kitchen scenarios to placement scores after optimization. We place the object at the new pose and then run 500 simulation steps to allow the object to settle into its final pose. We then compare the discriminator's confidence score with how much the object moved.
For these experiments we generated 100 random scenes, each with a random predicate goal so as not to bias it to a particular subset of the problem space. We ignore scenes if no feasible pose was found according to the discriminator. 

We found that in 95\% of these scenarios, the discriminator was able to find a stable pose to place a particular object, and in 90\% of all scenarios the planner was also able to match the specified predicates. This shows that not only can we find realistic positions, but that the discriminator does not preclude achieving specified goals. Note the higher perceived success rates than in the planner comparison in Table~\ref{table:planner}: this is because we only test scenes where the planner was initially confident that it could find a solution.

\begin{figure}[hbt]
\includegraphics[width=\textwidth]{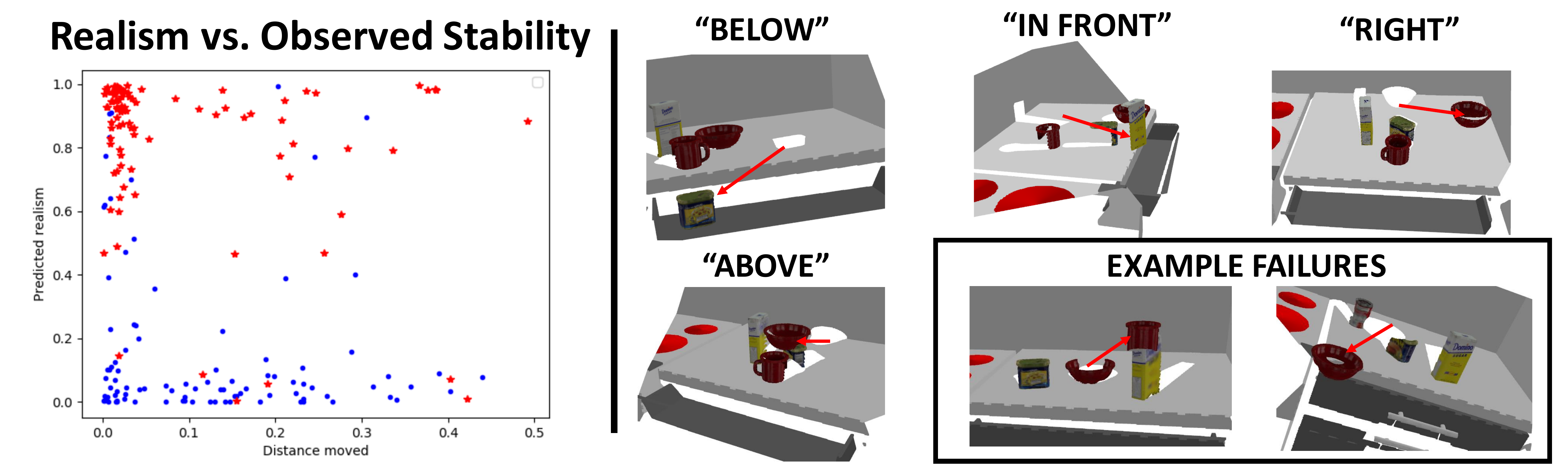}
\caption{Sim experiment results. \textbf{Left:} correlation between predicted realism (y axis) and distance moved after placement (x axis). The blue dots show randomly sampled poses, while the red stars indicate poses after running the placement planner. More realistic poses are much more stable than less realistic ones, although in some cases unrealistic poses do not move much either. \textbf{Right:} selected successful and unsuccessful simulation results. Images show point clouds from the robot's point of view, after planning.}
\label{fig:sim-experiments}
\end{figure}
Fig.~\ref{fig:sim-experiments}~(left) shows the relationship between the distance the object moved once placed and the output of the scene discriminator model, after rejecting a small number of outliers (distance $\ge$ 0.5 meters). When we compute Pearson's correlation coefficient $r$,
we see that for a single random sample $s$, $r_s = -0.147$ with $p_s=0.161$, and for the final predicted placement $P$, $r_P = -0.278$ with $p_P = 0.006$. Individual initial samples $s$ are shown in blue 
and optimized placements in red. These results indicate that our scene discriminator learns a metric which correlates with stability, even in challenging, cluttered scenes.
Failures of the discriminator are driven by oddly perceived object or environment geometry.
When parts of the object geometry are missing, the model is prone to making mistakes, such as placing an object so that it intersects with another object, or placing an object so that it is not properly supported, as in shown in Fig.~\ref{fig:sim-experiments}~(right).



\subsection{Predicate Performance}

\begin{table}[bt]
\centering \small
\begin{tabular}{l c c c c c c c c c}
    \toprule
    & Left of & Right of & In Front & Behind & Above & Below & Near & Touching & Centered \\
    \midrule
    Learned & 0.911 & \textbf{0.929} & \textbf{0.759} & 0.653 & \textbf{0.867} & \textbf{0.822} & \textbf{0.869} & \textbf{0.923} & \textbf{0.659} \\
    Baseline & \textbf{0.914} & 0.885 & 0.660 & \textbf{0.852} & 0.784 & 0.756 & 0.825 & 0.418 & 0.035 \\
    \midrule
    \%True & 13.9\% & 14.0\% & 4.6\% & 5.0\% & 4.3\% & 4.3\% & 29.0\% & 12.7\% & 5.6\% \\
    \%False & 81.6\% & 86.0\% & 95.4\% & 95.0\% & 95.7\% & 95.7\% & 71.0\% & 96.3\% & 84.4\% \\
    \bottomrule
\end{tabular}
\caption{\revised{F1-score of the predicate predictor $p_\rho$ in held-out randomly-generated simulated test scenes. Some predicates in our scenes can be very difficult due to clutter and occlusions, but our learned models are either on par with or better than almost all baselines. Bottom two rows show prevalence in the evaluation data set.}}
\label{tab:dataset-results}
\vskip -12pt
\end{table}

Table~\ref{tab:dataset-results} \revised{shows the F1 score by predicate} on a held-out portion of the dataset containing 2696 examples. Generally, our model classifies different relationships very well, matching or exceeding the baseline in every case but one. \revised{In particular, our learned model almost always outperforms the rule-based baseline in terms of \textit{sensitivity}, meaning that we can more often find valid positions for the object when attempting to find placement positions that satisfy the discriminator.}

\revised{By contrast, the rule-based approach is more likely to classify valid configurations as negative.} Certain predicates are much harder than others: these appear very rarely, such as \revised{\textit{behind} (objects are often far from the camera and multiply-occluded). Finally, the rule-based approach requires implementing and hand-tuning a range of different predicates; by learning from data, we could in principle scale to a larger number of symbolic relationships.}

\textbf{Baseline implementation}: First, we compute bounding boxes and means from each object point cloud. For the \textit{centered} predicate, we compute $xy$ distance on the table and use the same threshold that appears in our dataset (1 mm). For \textit{touching}, we compute minimum distance between point clouds and threshold it with 2.5mm. For \textit{near}, the threshold is a 5 cm distance. For directional predicates, we apply the exact same rules used in data generation to the computed bounding boxes\revised{, as seen in the supplementary materials.} 

\subsection{Real World Experiments}\label{sec:case_study}
Finally, we deployed our system on a Franka Panda robot with an arm-mounted RGB-D camera.
We used objects from the YCB object set~\cite{calli2015ycb} augmented with toy kitchen objects. We also used various cardboard boxes to force the robot to adapt to changes in scene geometry.
We generate grasps using~\cite{sundermeyer2021contactgraspnet} and use RRT-Connect~\cite{kuffner2000rrt} for motion planning. \revised{We used unseen object instance segmentation from~\cite{xiang2020learning} to determine segmentation masks for different objects, and use a prompt to choose which object to move when performing experiments.}
We compute standoff poses for each object 10cm above the predicted goal position, and release it 2cm above the predicted pose in order to ensure that we do not press into the table. Figure~\ref{fig:cover-pairs} shows example pairs of before-after images.





\revised{To quantify our results in the real world, we conducted a set of experiments with eight different objects. We report results in Table~\ref{tab:real-world}.}
\revised{
  Our placement planner was highly successful at finding manipulation plans with a range of different objects and predicates, including both grasps and placements for the various held-out objects.}
Crucially, our method is able to find multiple valid solutions for each scene; some examples appear in the supplementary materials. 
We saw a high rate of grasp execution failures on the real world, which could easily be improved in future work by re-grasping.


\begin{table*}[bht]
    \small
    \centering
    \begin{tabular}{l c c c c c c}
        \toprule
        & \multicolumn{3}{c}{Table Only}  & \multicolumn{3}{c}{Table with Large Box} \\
        Success Rate & Plan & Grasp & Placement & Plan & Grasp & Placement \\
        \midrule
        Sauce Bottle & 3 & 2 & 2 & 2 & 1 & 1 \\
        Cookie Box & 3 & 3 & 3 & 3 & 1 & 0 \\
        Red Mug & 3 & 3 & 3 & 3 & 2 & 1 \\
        Juice Carton & 3 & 0 & - & 1 & 1 & 1 \\
        Macaroni Box & 3 & 2 & 2 & 3 & 0 & - \\
        Parmesan Can & 3 & 0 & - & 1 & 0 & - \\
        Mustard Bottle & 3 & 3 & 3 & 2 & 2 & 2 \\
        Large Pitcher & 2 & 2 & 2 & 3 & 2 & 0 \\
        \midrule
        Overall (\%) & 95\% & 75\% & 100\% & 75\% & 56\% & 60\% \\
        \bottomrule
    \end{tabular}
    \vskip 6pt
    \caption{Generalization experiments with different objects. All numbers out of three trials. Placement successes conditioned on successful grasps. We see that the largest cause of failures was grasping issues. Planning failed in a few situations with challenging objects that were either very large (YCB pitcher) or very small and hard to grasp (Parmesan can). Placement fails when an object falls off of the box after the planner attempts placement.} 
    \label{tab:real-world}
    \vskip -18pt
\end{table*}


\section{Conclusions and Future Work}\label{sec:conclusion}
We demonstrated the ability for a robot to learn to accurately infer
inter-object relations from point clouds of real-world, unstructured
environments, which can be used to perform manipulation planning for previously unseen objects. 
Our results show that a model trained purely in simulation, effectively predicts relations on real-world point clouds of objects not seen during training. Furthermore, our incorporation of a scene-realism discriminator significantly improves performance over the predicate goal predictor alone.  
In the future, we will add semantic understanding about which objects the robot is interacting with using learned object class and attribute classifiers. We will also expand our method to handle multi-object relations and explore long-horizon manipulation tasks by formally extending our method into a task and motion planner. 



\bibliography{main}

\clearpage
\newpage
\appendix
\section*{Supplemental Materials}

We provide two appendices. Appendix~\ref{sec:details} includes a detailed explanation of our predicates, model architectures, and training. Appendix~\ref{sec:extra-real-world} contains additional qualitative results, including a number of real world placement examples and a discussion of our results.

\section{Implementation Details}\label{sec:details}

This section describes how we created our dataset and trained the model. One major advantage of our approach is that it only relies on having positive data; we automatically generate ``unrealistic'' data during the training process with which to train our scene-realism predictor.

\subsection{Dataset Implementation}

Each scene consists of 3 to 7 random Shapenet~\cite{shapenet2015} objects in stable configurations on various surfaces, including in stacks. We include mugs, bowls, plates, bottles, pots, and various small objects, as well as boxes and cylinders of random sizes. We specifically included only objects that appear in the ACRONYM grasp dataset~\cite{acronym2020} as well as in Shapenet~\cite{shapenet2015}.

\begin{figure}[bht]
\centering
\includegraphics[width=\textwidth]{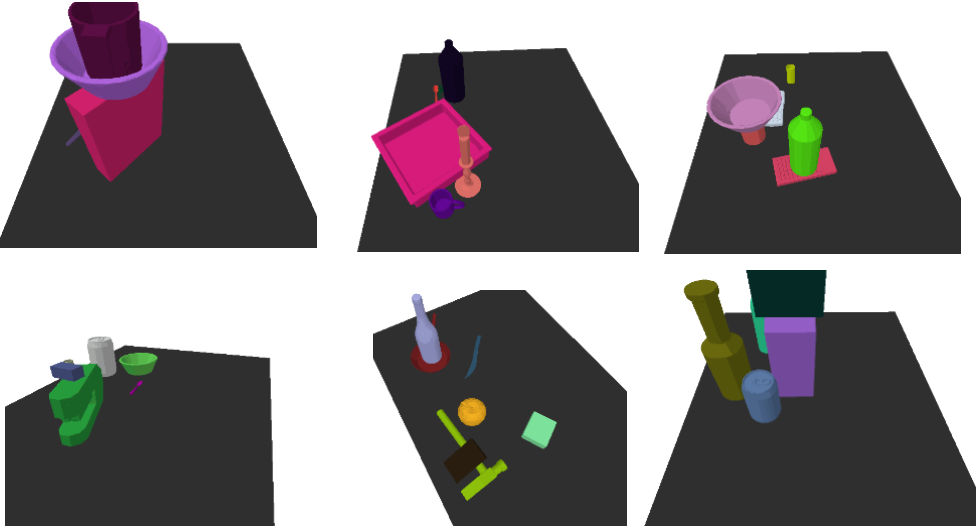}
\caption{Examples of PyBullet scenes containing objects placed in different locations. We train on these scenes, with each object either hidden if it is query (to get scene and anchor embeddings) or hidden if it is \textit{not} query (to get the query object embedding).
}
\label{fig:data-example}
\end{figure}


We use PyBullet\footnote{\url{https://pybullet.org}} to simulate scenes. 
\revised{Figure~\ref{fig:data-example} includes renderings of some example scenes containing the shapenet objects.}
Objects are moved around throughout the scene to generate data. 
In order to train the rotation model, we take the point cloud from before the motion, apply the correct rotation, and move it into its ``observed'' position. This constitutes a positive example that can be used to train our placement planner.
Objects are placed either in a random orientation at a random height above the planar surface, or on top of a flat surface (which can include the table or other objects, inducing object stacks). The physics engine is then simulated forward until the objects come to rest, which results in a stable scene.

We used 5563 scenes for training our orientation model and 427 scenes for testing it. These scenes are shown in the supplemental video.
In practice, we also tested a version of our model that was trained purely on static scenes, without orientation, and with only the directional predicates. 
This dataset contained generated 25,441 scenes with 5 images per scene from random viewpoints, resulting in a total of 221,336 relational predicates. Additional simulation results in Sec.~\ref{sec:extra-sim} are using this model. 

We define the following predicates in our simulator: \texttt{above}, \texttt{left_of}, \texttt{right_of}, \texttt{touching}, \texttt{below}, \texttt{near}, \texttt{aligned}, \texttt{centered}, \texttt{behind}, and \texttt{in_front_of}.

\subsubsection{Directional Predicate Implementation} \label{sec:directional}
We define the predicates shown in Table~\ref{tab:dataset-results}.
Let $i, j$ denote the objects that will be considered for a pairwise predicate, and $o_i$ be the 3D center of object $i$. We first obtain the eight bounding box corners and center for $i, j$ in the camera frame. We use a left-handed coordinate system with the x-axis pointing right, y-axis pointing up. For the z-axis, we have it pointing towards the scene (away from the user), but we orient it such that it is parallel to the planar surface so that the above/below predicates are easy to compute.

\begin{wrapfigure}{r}{0.3\textwidth}
 {\centering
    \vspace{-12pt}
\includegraphics[width=\linewidth]{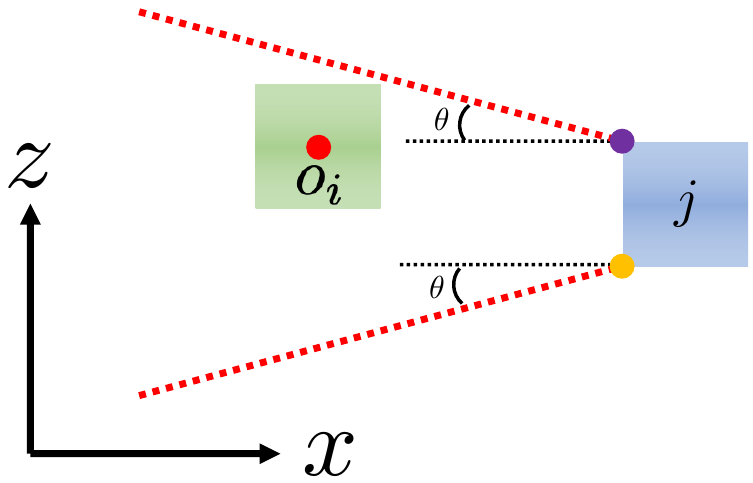}
\caption{Illustration of \textbf{left} predicate for the $xz$-plane.}
\label{fig:left_right}
\vspace{-12pt}
}
\end{wrapfigure}

Here, we give a high-level description of the algorithm used to determine the \textbf{left} predicate (i.e. is $i$ to the \textbf{left} of $j$?).
Essentially, we compute whether $o_i$ is within the trapezoidal volume defined by $j$'s bounding box corners, and an angle $\theta$. To implement this, we can first check this in the $xz$-plane. Figure~\ref{fig:left_right} shows a mock-up of this in the $xz$-plane.
Object $i$'s center (red dot in green object) must lie below the line defined by object $j$'s upper corner (purple dot) and $\theta$,
which is denoted with the red dashed line.
Similarly, it must lie above the line defined by the bottom corner (yellow dot) and $\theta$.
Lastly, $o_i$ must be to the left of $j$'s bounding box corners, which completes the trapezoidal volume definition in the $xz$-plane. This same computation is repeated for the $xy$-plane to obtain the 3D trapezoidal volume. Note that this computation assumes full knowledge of the object bounding boxes and centers, which we can obtain via the simulator.

Specifically, we consider the following set of rules:
\begin{enumerate}
    \item $o_i$ must be in the half-space defined by o2 upper corner and $\theta$ in the $xz$-plane
    \item $o_i$ must be in the half-space defined by o2 lower corner and $\theta$ in the $xz$-plane
    \item do same as 1) for $xy$-plane
    \item do same as 2) for $xy$-plane
    \item $i$'s center must be to left of all o2 corners
    \item All corners of object $i$ must be to the left of $j$'s center
\end{enumerate}

\textbf{Right} can be computed by applying the same set of rules to the flipped order of objects: $j, i$. Additionally, \textbf{front/behind} and \textbf{above/below} can be computed with the exact same set of rules, but considering different planes (e.g. $xy$ and $zy$ planes for \textbf{above/below}).


\textbf{Model-based baseline}: on point cloud data, we simply use the bounding box computed from the labeled point cloud for each object as a model-based baseline to compare against. We apply the exact same rules as in the point cloud cases.

\subsection{Other Predicates}

We have four additional predicates in our dataset: \texttt{centered}, \texttt{touching}, \texttt{near}, and \texttt{aligned}. Both \texttt{touching} and \texttt{near} are defined based on mesh geometry. The mesh distance thresholds are set to 1mm for \texttt{touching} and 5 cm for \texttt{near}, but are based on the actual mesh as it appears in each scene.

The \texttt{centered} and \texttt{aligned} predicates are both based on the ground-truth pose of the objects. \texttt{centered} is true if the center of the object is object is within 1 mm. Two objects were considered \texttt{aligned} if the difference in orientation was less than $\pi/20$. In practice, our dataset was filled with Shapenet objects~\cite{shapenet2015}, which were not well aligned, so this proved difficult to learn; this can be fixed with better object annotations and will be explored in depth in future work.

\textbf{Model-based baseline}: We compute a reasonable model-based baseline for each approach in our dataset. For the \texttt{centered} predicate, we compute $xy$ distance on the table and use the same threshold that appears in our dataset (1 mm). For \texttt{touching}, we compute minimum distance between point clouds and threshold it with 2.5mm. For \texttt{near}, the threshold is a 5 cm distance.


\subsection{Model Architecture}
\label{sec:model-details}

All pointnets are trained just on $xyz$ for each point.
Our encoder $e(x)$ is a Pointnet++ network with three set abstraction layers. These are:
\begin{enumerate}
    \item 128 points, 64 samples, scale of 0.04, and an MLP of size [3, 32, 32, 64]
    \item 32 points, 32 samples, scale of 0.08, and MLP of size [64, 64, 64, 128]
    \item A full set Pointnet layer with MLP of [128, 128, 128, 256]
\end{enumerate}

This is followed by a single fully connected layer going from the input size to 256 with ReLU and layer norm, and then down to an $h$ of size 128. Input size is the PointNet encoder output, concatenated with a 128-D encoding of the object center output by a second MLP.

We concatenate $h_i$ and $h_j$ as inputs to the predicate classifier $p_\rho(h_i, h_j)$.
It consists of a single fully connected layer going from 256 to 128, and then to the number of predicates (nine, in our case; one was omitted because it was not applicable to real world experiments).

The prior $\pi$ takes the concatenation of $h_Z, h_i, h_j$, and the predicate goal $\vec{\rho}$. It is then a fully connected layer from $384 + N_{predicates}$ to 128, followed by a batchnorm. This then goes into the MDN, which predicted $k=5$ clusters. $\delta$ is the 3D vector $xyz$.

The discriminator was a 4-layer Pointnet, with four set abstraction layers:
\begin{enumerate}
    \item 512 points, 64 samples, scale of 0.05, and an MLP of size [3, 32, 64]
    \item 256 points, 64 samples, scale of 0.10, and MLP of size [64, 64, 128]
    \item 128 points, 64 samples, scale of 0.20, and MLP of size [128, 128, 256]
    \item A full set Pointnet layer with MLP of [256, 256, 512]
\end{enumerate}
The layer ends in a norm and a ReLU activation, a fully connected layer, and maps to a 1d output with sigmoid activation.
All models were implemented using Pytorch Pointnet++~\cite{pytorchpointnet++}.

\subsection{Rotation Training}

We trained a version of the model that predicts orientation instead of just $xyz$ Cartesian transformations, as from previous work (e.g.~\cite{qureshi2021nerp}). However, training the rotation prior requires a slightly different consideration from Cartesian motions. We take two subsequent frames in our rotation placement dataset and use the frame at time $t-1$ for an observation at time $t$ to extract the object's point cloud. We compute the rotation $\theta$ as the component of the transformation around the world's z axis and use this rotation from $t-1$ to $t$ to train the rotational component of our MDN prior.

When training models on data without rotations, we can instead use the frame at time $t$ alone. Instead, we save extra images with and without each object rendered in the scene, to get virtual placement data.

\subsection{Segmentation and Grasping}

\begin{figure}[bt]
\centering
\includegraphics[width=0.49\textwidth]{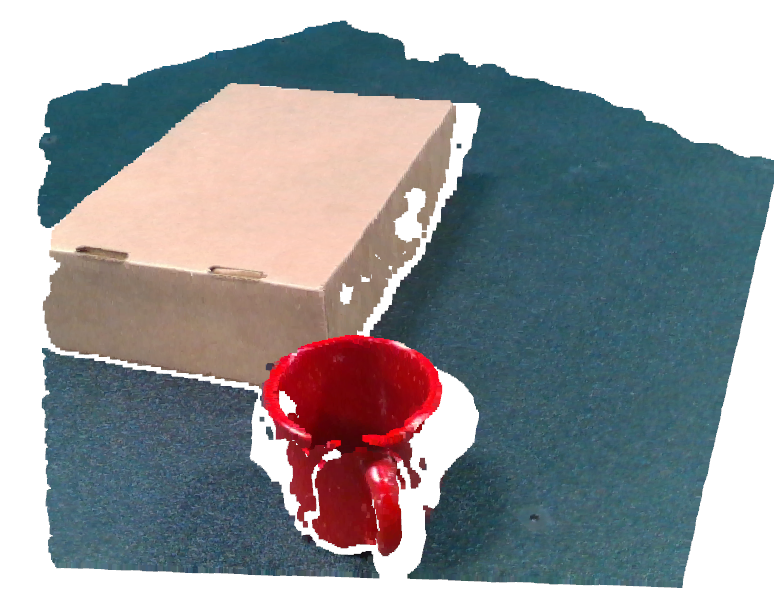}
\includegraphics[width=0.49\textwidth]{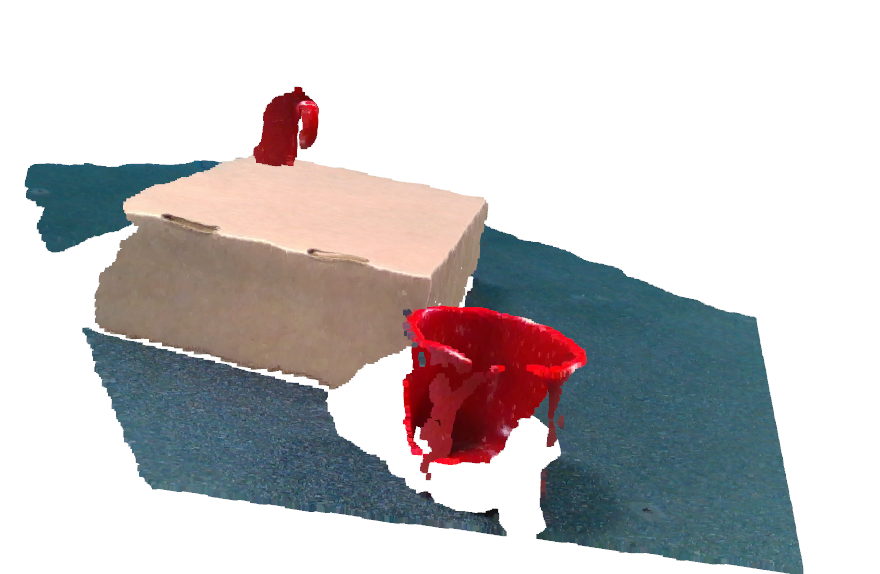}
\vskip 12pt
\caption{
Example of segmentation causing issues when manipulating objects. In this case, the mug was broken up into two objects. Though the object was correctly placed, this made grasping very difficult because the available grasps made less sense.}
\label{fig:segmentation-issues}
\end{figure}

We used recent work by Xiang et al.~\cite{xiang2020learning} for unknown object instance segmentation, as the code is open source and available on Github\footnote{\url{https://github.com/NVlabs/UnseenObjectClustering}}. This worked very well in many cases, but needed some minor modifications in order to be used on the real world.

In particular, we had a problem with individual pixels from around the objects being labeled as a part of those objects. This meant that even after moving an object, parts of it would be left behind, creating geometry that the planner could occasionally place things on top of. We mitigated this problem by adding a dilate and erode to each object mask, and -- crucially -- ignoring any 3d points that fell within the region of this dilation and erosion. This is the source of the white boundaries around all of our objects in these figures.

Even with these modifications we still saw occasional issues where segmentation problems could cause planning issues;
see Fig.~\ref{fig:segmentation-issues} for an example. In this case, the red mug - an object that was generally very easy to reliably grasp during our experiments -- was broken up into two different pieces before being placed ``above'' the large cardboard box. While the placement planning component was successful, this has clear ramifications for grasping -- in particular, the generated grasps are of lower quality, and often in cases like this our motion planning will fail for safety reasons.

In the real world, grasps were computed via 6-DOF Graspnet~\cite{mousavian20196}, which gives us a range of grasp poses that we can use for each object. Many failures we observed in practice were due to unpredictable behavior of objects once grasped, or due to the fact that grasps would fail.


\section{Additional Experiments}\label{sec:extra-real-world}

\begin{figure}[h!]
\centering
\includegraphics[width=0.8\textwidth]{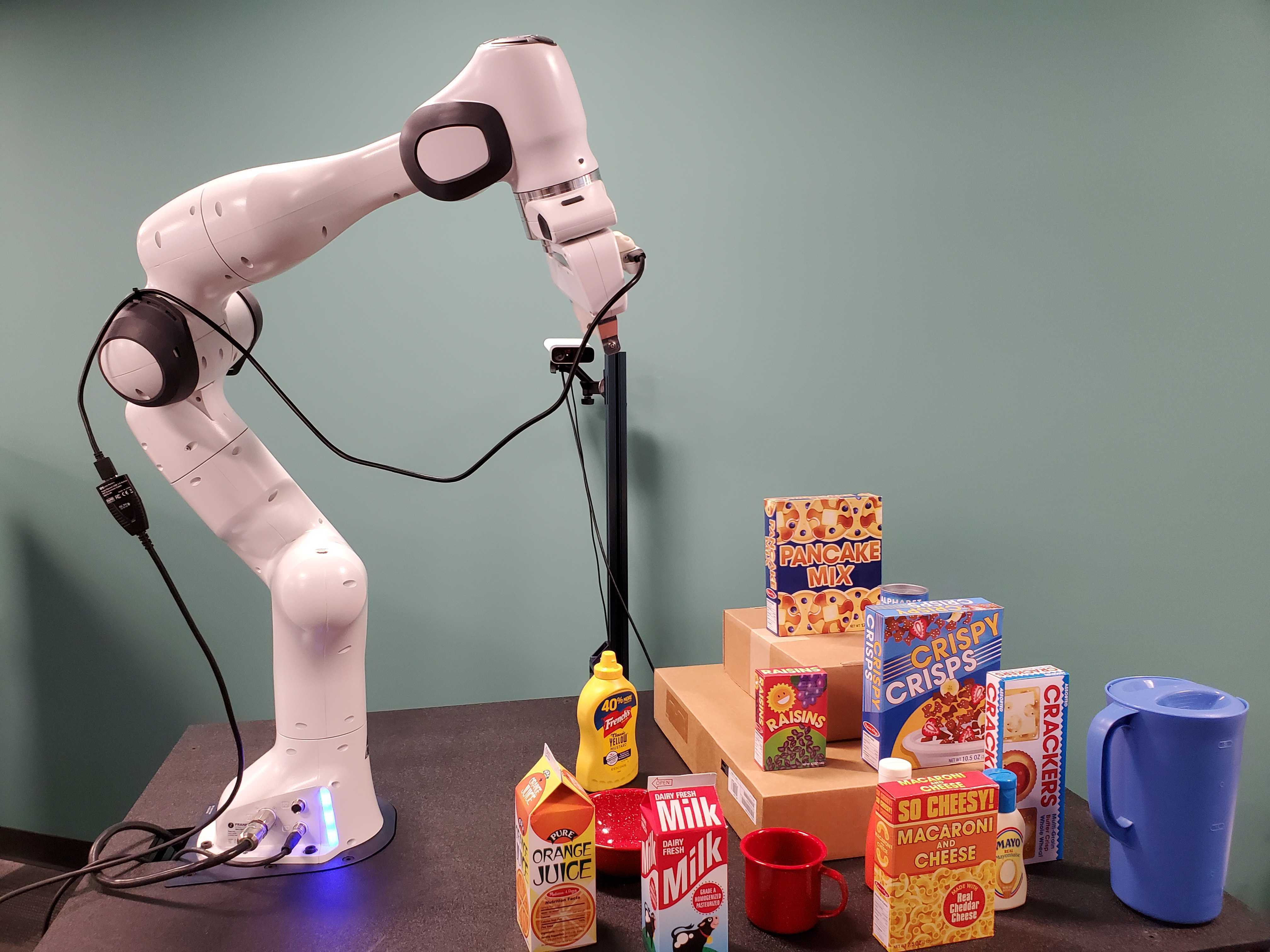}
\vskip 12pt
\caption{Real-world setup with multiple objects. The robot has an Intel Realsense D415 RGB-D camera mounted on its end effector at a known offset. We changed the scene geometry by positioning several large boxes.}
\label{fig:robot-setup}
\end{figure}

We implemented our system on a real-world robot rearrangement task and show a variety of results for how our placement planner works in practice.
Figure~\ref{fig:robot-setup} shows the real-world scenario, with the robot and a selection of the objects we tested on. We used ROS\footnote{\url{https://ros.org}} for messaging and execution~\cite{quigley2009ros}.
The setup also has a Microsoft Azure camera for viewing the entire scene, but this was not used in these experiments; instead, we used an Intel Realsense D415 camera to capture RGB-D images of the scene.

All tests were performed with images taken from the same camera pose to the right side of the robot, and then executed closed-loop after a manipulation plan was found.
We plan motions using RRT-Connect~\cite{kuffner2000rrt}.

\textbf{Object selection.} Objects were selected via a user interface, where the user would input the number associated with a particular mask (0 through $N$) for each part of the query.

\subsection{Real World Placement}

\begin{figure}[h!]
\centering
\includegraphics[width=0.4\textwidth]{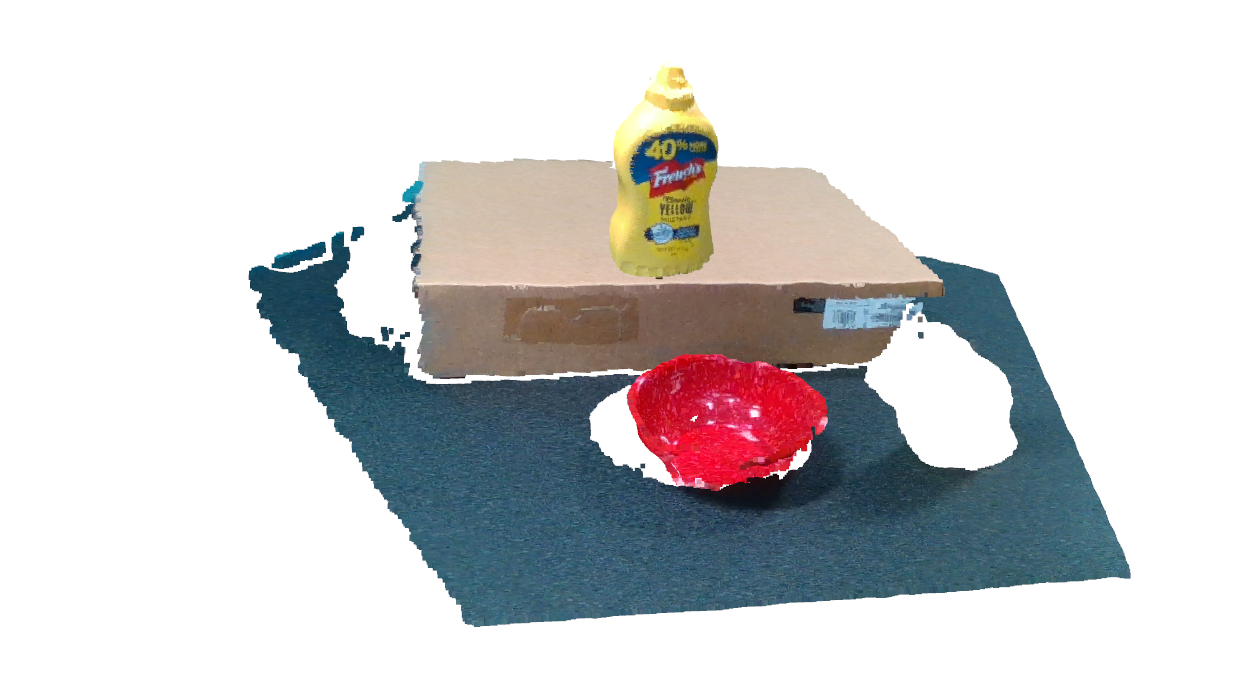}
\includegraphics[width=0.4\textwidth]{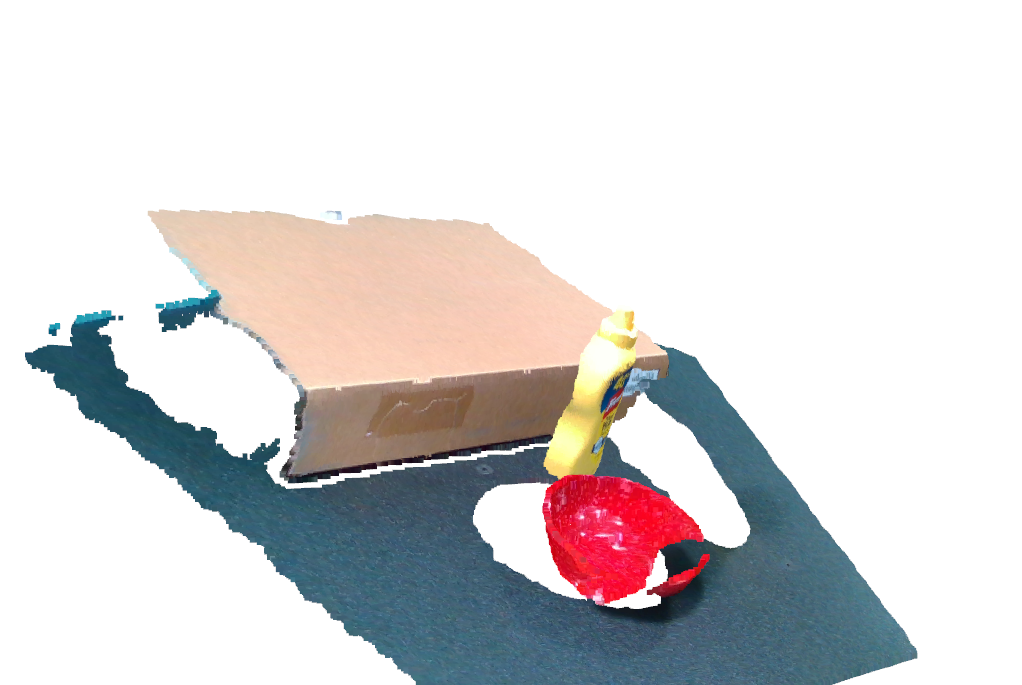}
\caption{
Placing mustard behind the bowl, with a large box in the way. There are two possible solutions to the problem: either placing the mustard on the table or placing it on top of the box. Our approach found stable placement positions in both locations.
}
\label{fig:box-place-mustard}
\vskip -12pt
\end{figure}

Fig.~\ref{fig:box-place-mustard} shows two examples of the model generalizing to a configuration where a large box obstructs much of the area. One major advantage of our approach is the ability to come up with a range of different placements satisfying various conditions.

\begin{figure}[h]
\centering
\includegraphics[width=0.24\textwidth]{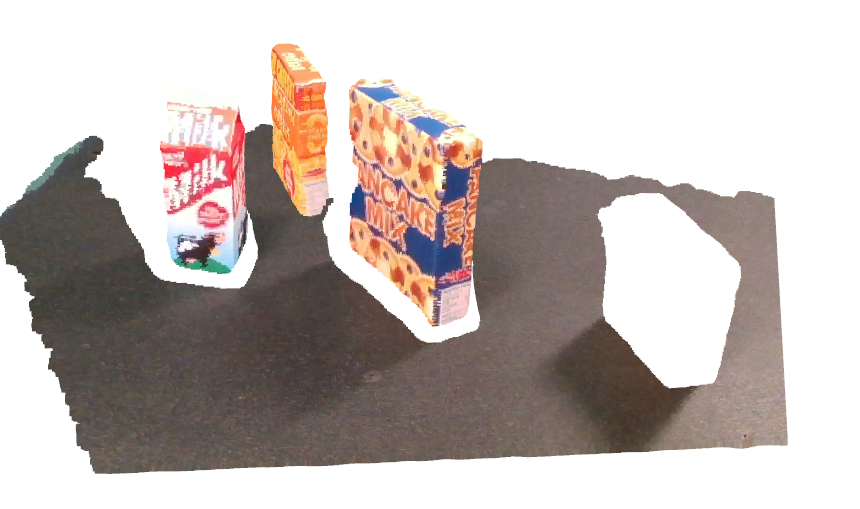}
\includegraphics[width=0.24\textwidth]{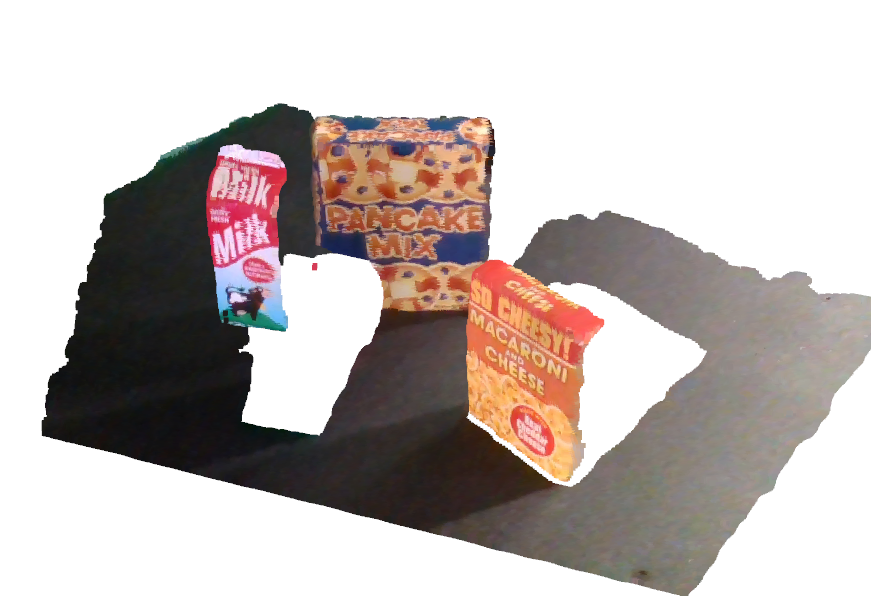}
\includegraphics[width=0.24\textwidth]{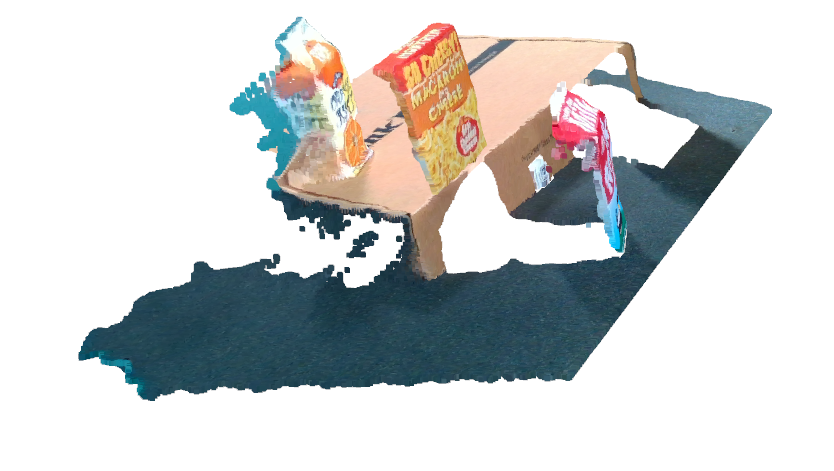}
\includegraphics[width=0.24\textwidth]{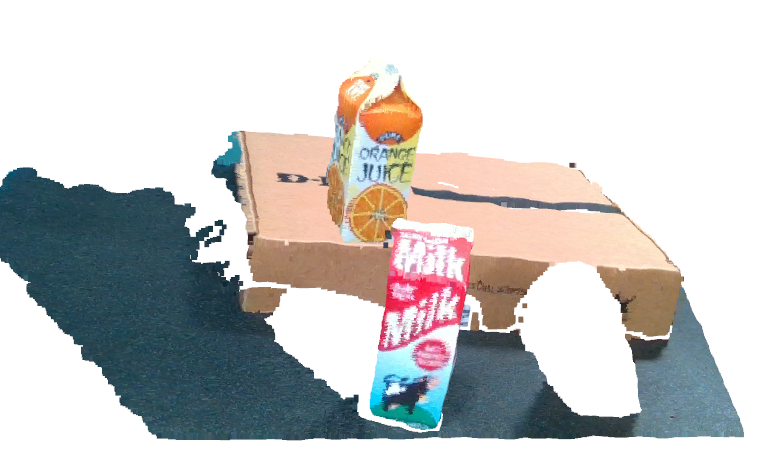}
\caption{Examples of placement actions given individual snapshots of real-world scenes. In the top row, the robot was asked to move either the orange macaroni box to behind the milk (left), or the milk to the left of the macaroni (right), on a flat table. In the bottom row, the goal was to place macaroni (left) or juice (right) behind the milk; we added a cardboard box, which means that the object has to be placed off the table.
}
\label{fig:real-world-examples}
\end{figure}

We changed the scene by adding various boxes and moving things around.
In one case, the system decides to place the mustard in front of the box, on the black cart surface; in the other, the mustard is placed on top of the box.
Fig.~\ref{fig:real-world-examples} shows some examples of different problems solved by our method. 

\begin{figure}[h!]
\centering
\includegraphics[width=0.24\textwidth]{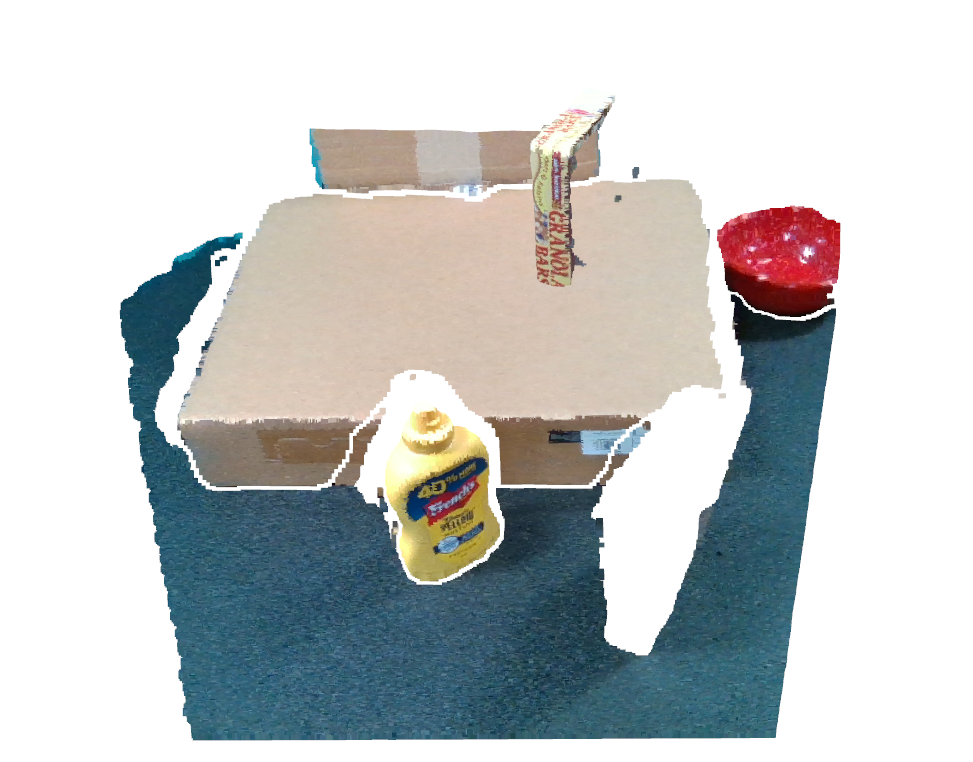}
\includegraphics[width=0.24\textwidth]{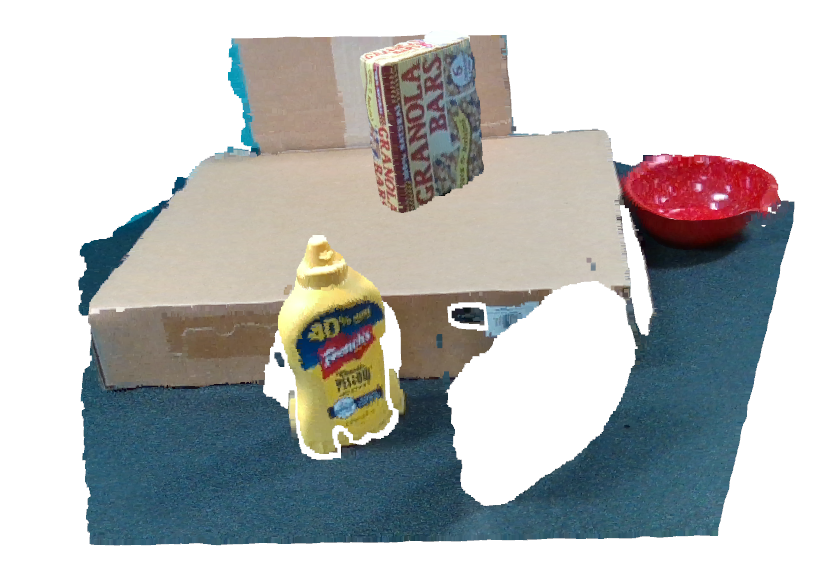}
\includegraphics[width=0.24\textwidth]{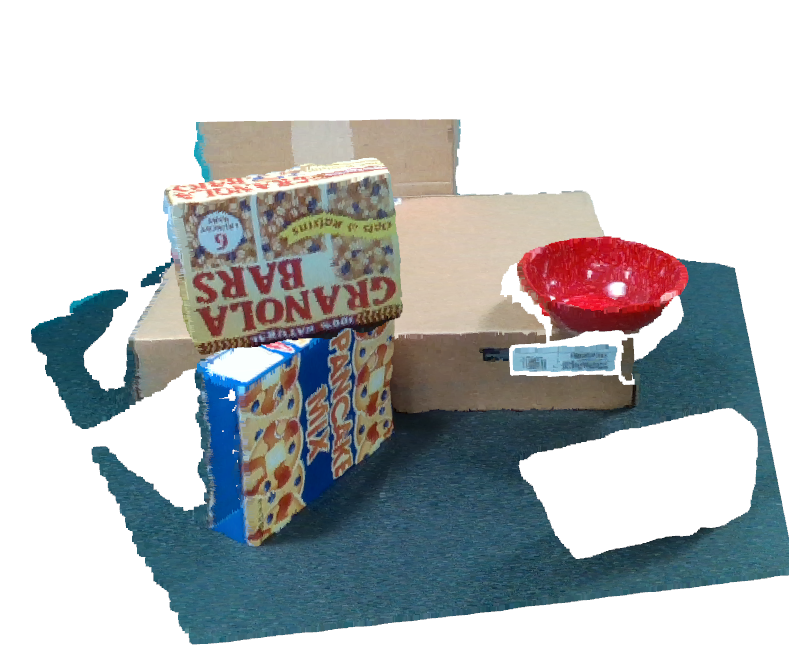}
\includegraphics[width=0.24\textwidth]{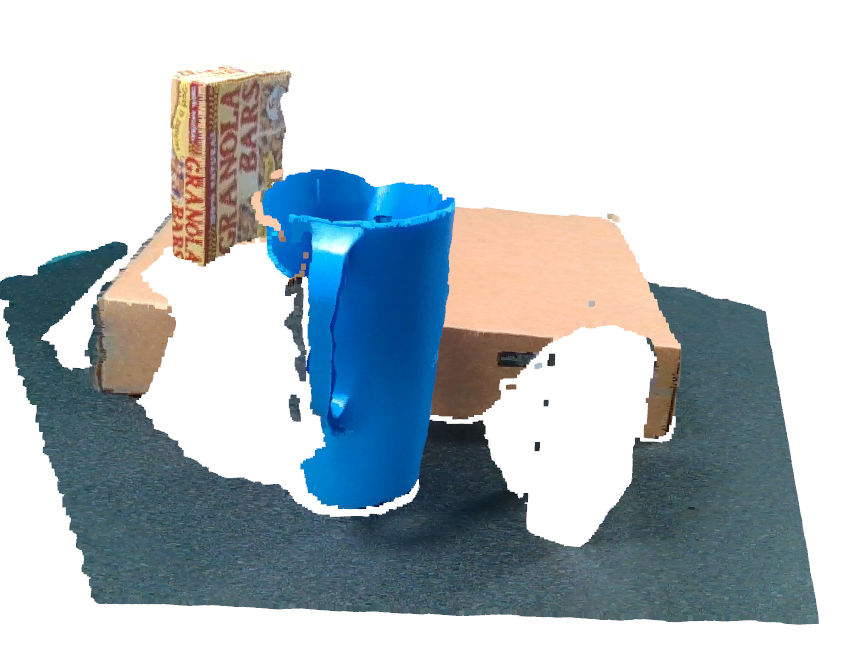}
\vskip 12pt
\caption{Some additional arrangement results for the ``granola box'' object. The planner finds a variety of poses, including stacking objects on top of one another, should that satisfy the specified predicate.}
\label{fig:extra-granola}
\end{figure}

Fig.~\ref{fig:extra-granola} shows some extra results for moving a box of granola around, placing it in different positions according to various predicate queries. Our discriminator $f$ is very capable of finding realistic-looking poses for a variety of objects; we see that it is able to plan a placement on top of the large cookie box rather easily, for example. In cases like this, execution is the main limitation preventing successes.

\begin{figure}[h!]
\centering
\includegraphics[width=\textwidth]{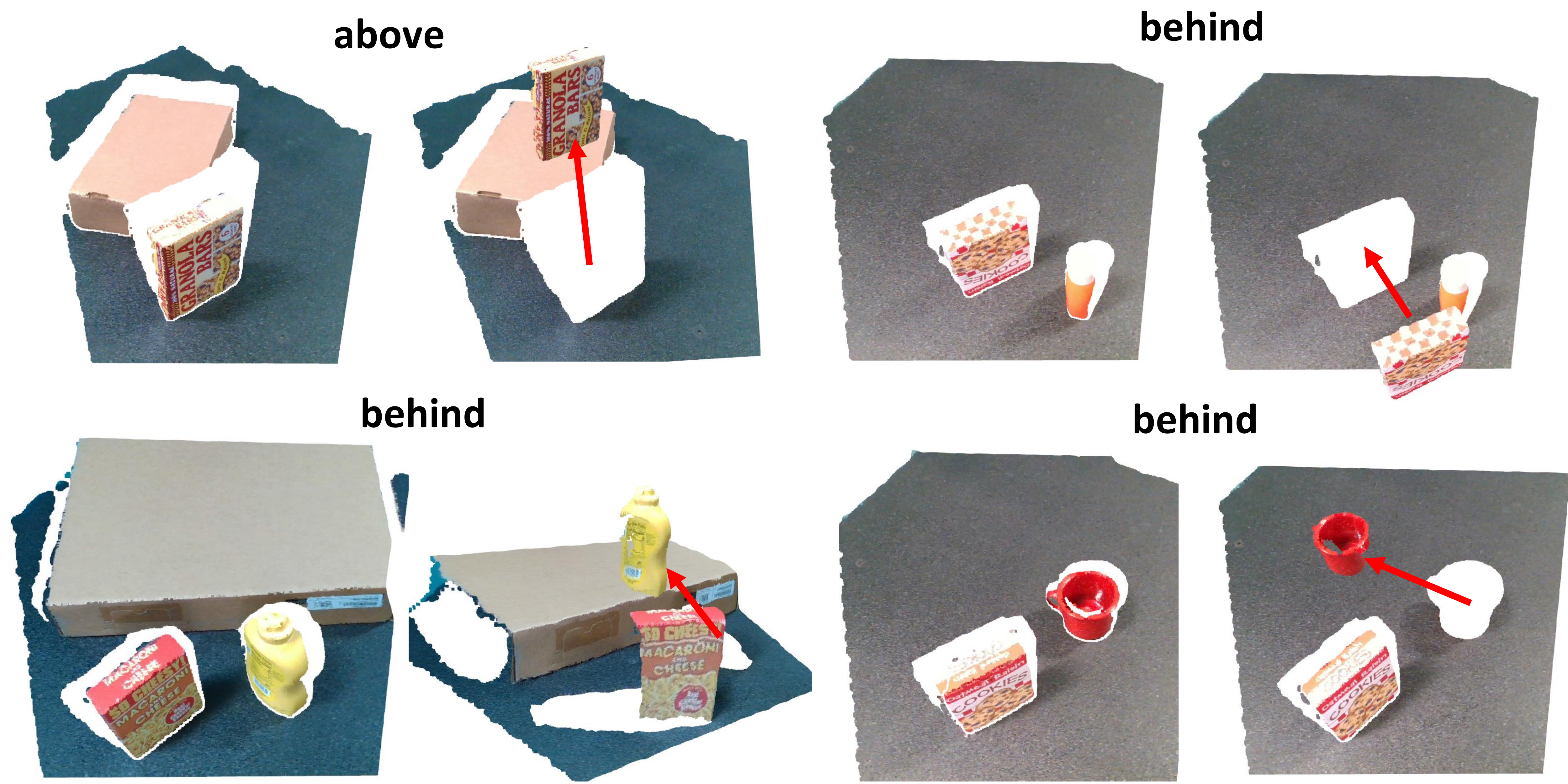}
\caption{Examples of before and after computation of final goal positions by our planner. For each pair, on the left, we show the initial scene observation, as read in by our planning algorithm. On the right, we show the planned goal position output by the model. The white gap denotes empty space where the object was moved from.}
\label{fig:extra-pairs-v1}
\end{figure}

There are white gaps left in each point cloud where the granola box was moved from. These are also visible in Fig.~\ref{fig:extra-pairs-v1}, which show a number of additional successful planning queries in a very simple environment.

\begin{figure}[h]
\centering
\includegraphics[width=\textwidth]{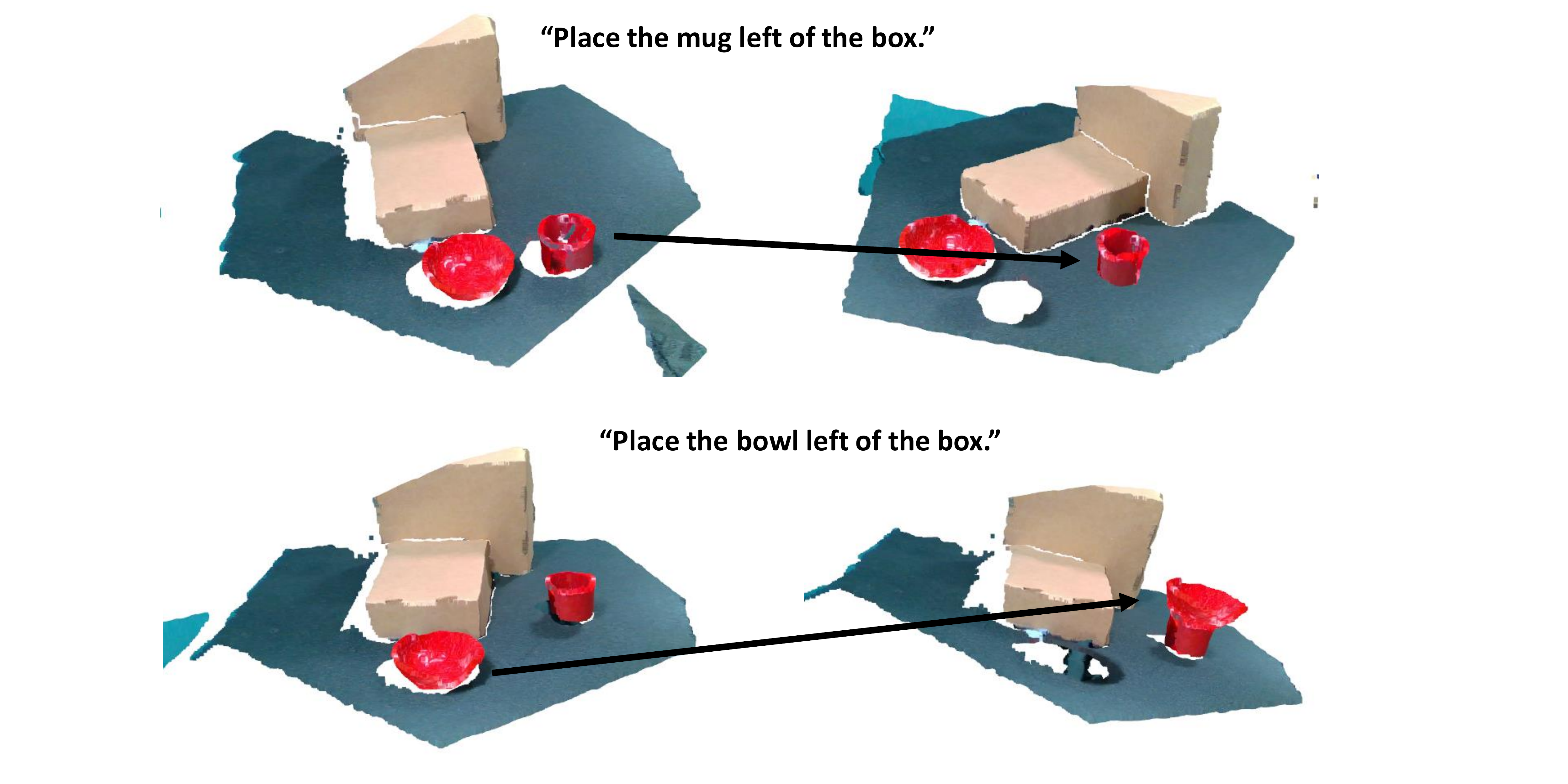}
\caption{A sequential manipulation. The system decides to place the bowl on top of the mug in the second step, as this is a valid placement that satisfies the specified goal condition without any collisions.}
\label{fig:mug-bowl-sequence}
\end{figure}

Sequential manipulation results can lead to interesting outcomes. In Fig.~\ref{fig:mug-bowl-sequence}, we see the results of performing a sequential manipulation, where the goal is to place the mug and bowl to the right of a large cardboard box. These sorts of actions are totally valid outcomes for our planner, and show how it has learned a variety of valid positions. Rather than picking something unrealistically close to the mug, it decides the safest thing to do is to simply stack the two.

Finally, we performed some tests in more complex or cluttered scenes, including multiple objects of which some are obstacles.

\begin{figure}[h]
\centering
\includegraphics[width=\textwidth]{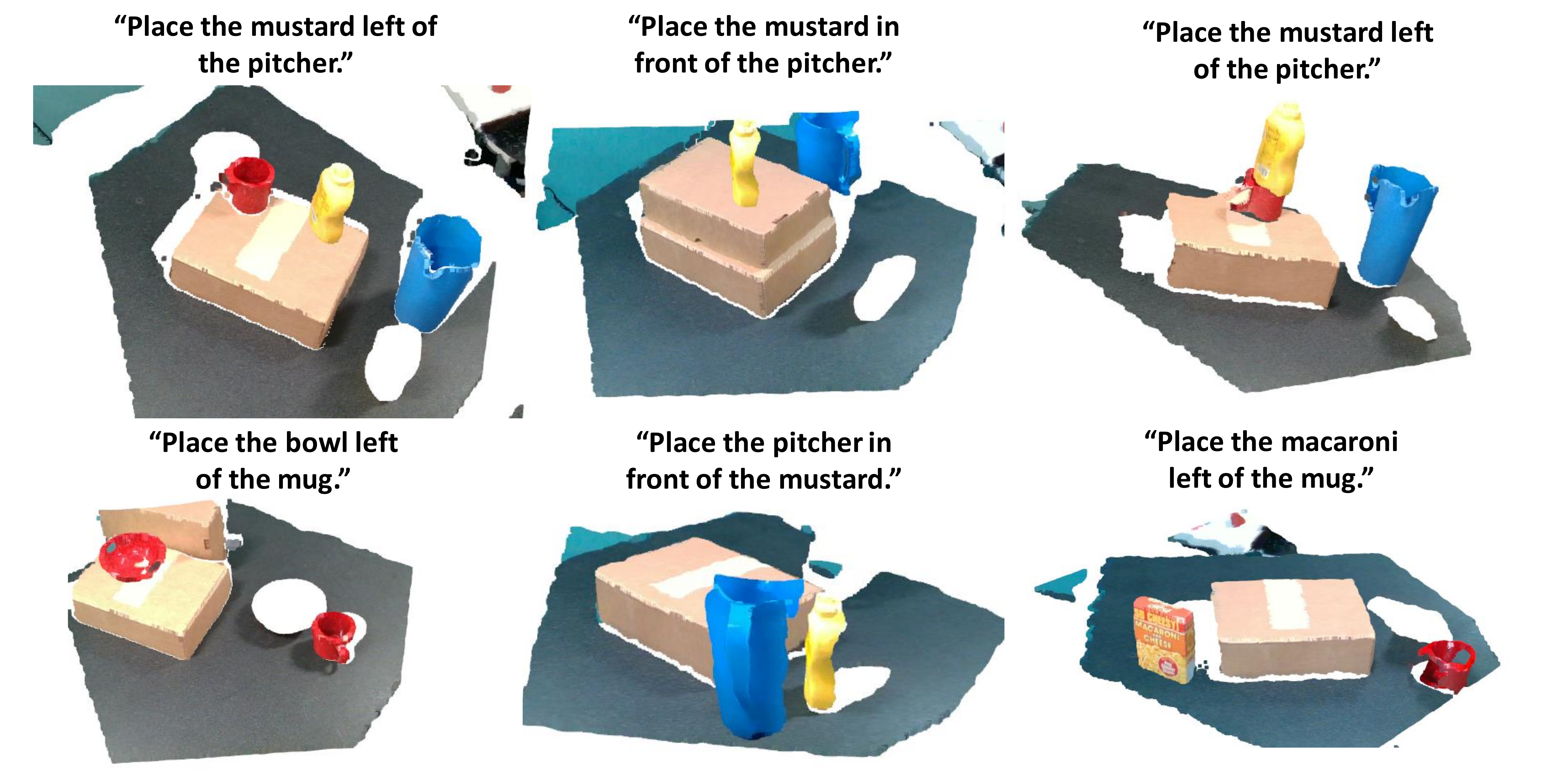}
\caption{A variety of objects moved into various configurations on the table, with different heights of obstacles and different predicates chosen as commands.}
\label{fig:extra-predicates-mustard}
\end{figure}

Fig.~\ref{fig:extra-predicates-mustard} includes a set of experiments performed with the mustard, where the planner must adapt to the mustard bottle while dealing with a mug that might be in several different positions or with a much higher placement surface. Depending on circumstances, the planner either attempts to place the object beside or on top of the mug.

We notice here that the discriminator generally avoids placing \textit{near} something unless explicitly told to; this is possibly a side effect of the discriminator training process. Due to differences between the simulated and real object dataset, the mustard placement in the top left of Fig.~\ref{fig:extra-predicates-mustard} is highly likely to fail, either falling into the mug or off of it.



\subsection{Cabinet Placement}\label{sec:extra-sim}

\begin{figure}[h]
\centering
\includegraphics[width=0.24\textwidth]{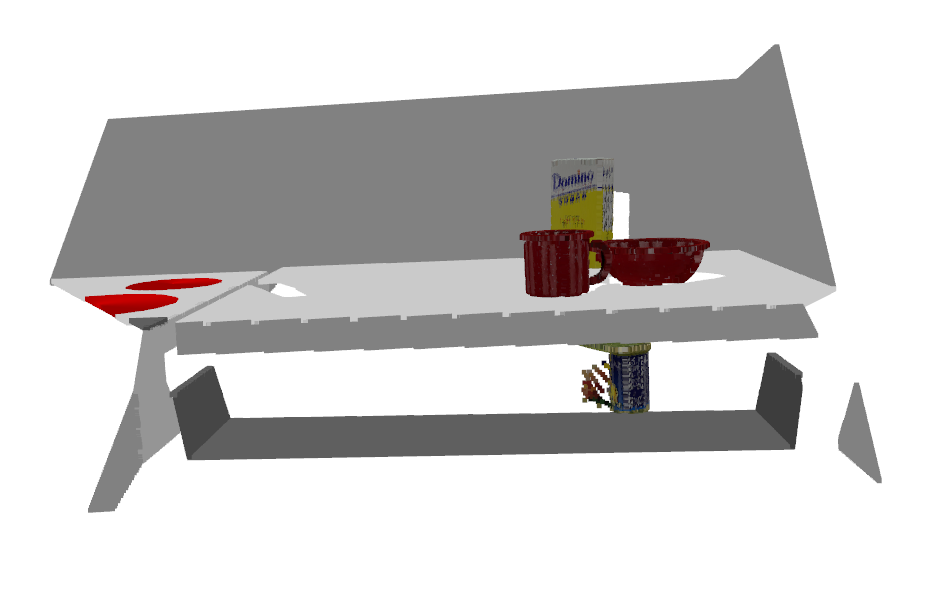}
\includegraphics[width=0.24\textwidth]{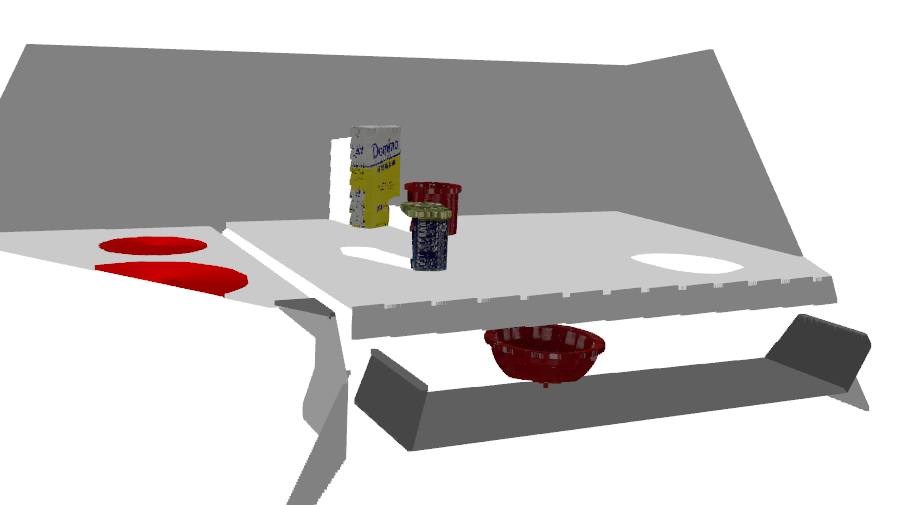}
\includegraphics[width=0.24\textwidth]{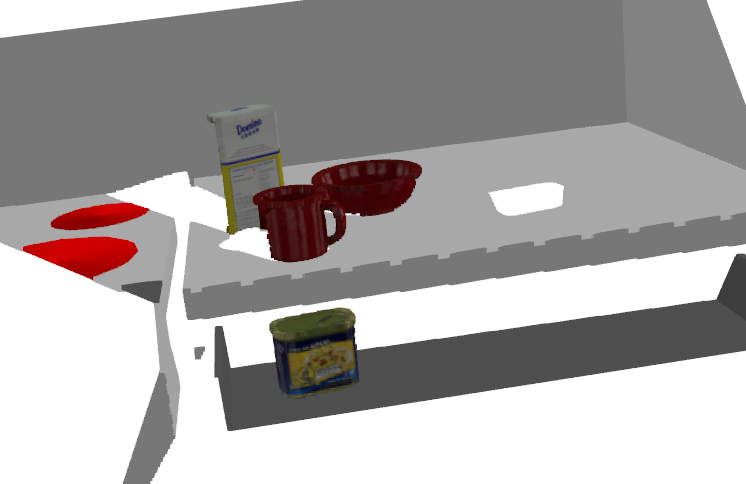}
\includegraphics[width=0.24\textwidth]{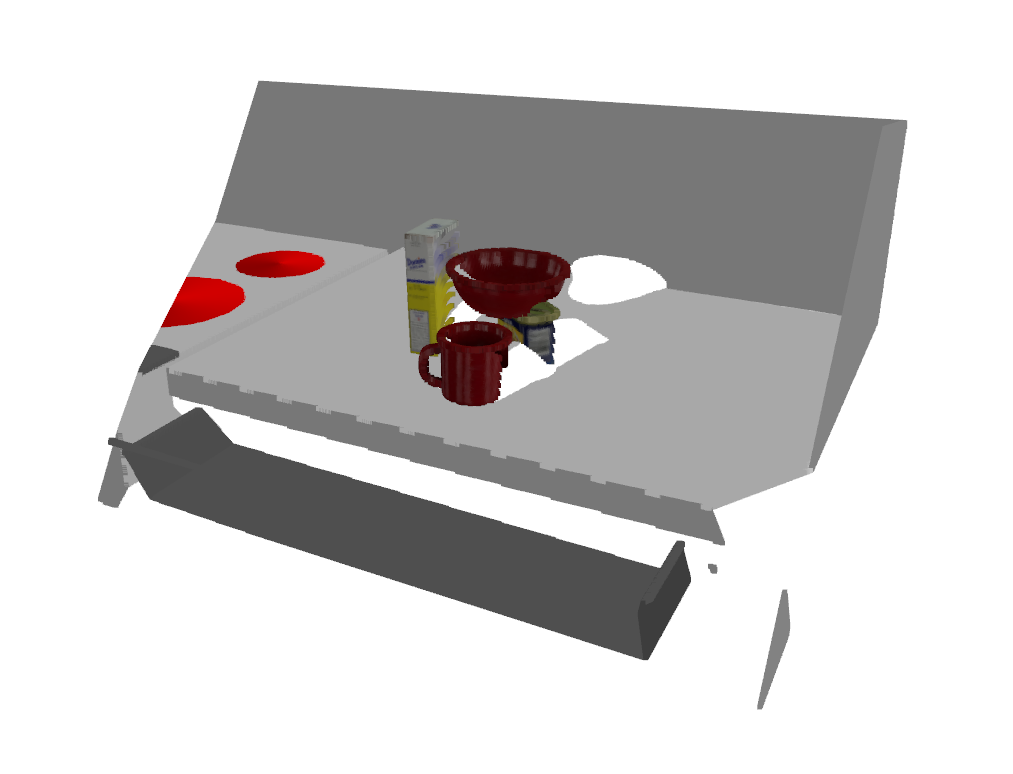}
\caption{
Examples of interesting placements in sim. The planner is able to place objects in the drawer (if told to place below a different object) or on top of the other objects, as is appropriate.
}
\label{fig:extra-sim-placements}
\end{figure}

Fig.~\ref{fig:extra-sim-placements} shows some additional placement results from simulation testing. These include placing the object in drawers, as appropriate, and stacking objects on top of one another, as in the lower right corner of the figure. The advantage of our model is how few assumptions it makes about the environment, which means that we can apply it to many different scenarios, even ones that we have not seen before.

\begin{table}[bt]
    \centering
    \begin{tabular}{l c c } 
        \toprule
        Predicate &  Sensitivity (\%) & Specificity (\%) \\ 
        \midrule
        Front & 98.8\% & 89.0\%\\
        Back & 96.2\% & 93.1\% \\
        Left & 96.2\% & 85.0\% \\
        Right &  96.5\% & 86.3\% \\
        Above & 79.3\% & 97.4\% \\
        Below & 79.3\% & 98.1\% \\
        \bottomrule
    \end{tabular}
    \vskip 6pt
    \caption{Accuracy of the predicate predictor $p_\rho$ in randomly-generated simulated test scenes by predicate.}
    
    \label{tab:dataset-results}
\end{table}

Table~\ref{tab:dataset-results} shows sensitivity (true positive rate) and specificity (true negative rate) by predicate on a set of randomly-generated simulation scenes, similar to Fig.~\ref{fig:extra-sim-placements}. To compute classifier accuracy, we generated $100$ random scenes and evaluated the classifier on each to determine if it was correct. To determine prior and planner accuracy, we sampled $100$ poses for each object and determined the accuracy of each pose. The hardest predicates to classify were those based on occlusions, presumably because 3d representations are not very useful for this.

\subsection{Additional Predicate Comparisons}

\begin{table}[bt]
    \centering
    \small
    \begin{tabular}{l c c c c c c c c } 
        \toprule
        Predicate & \multicolumn{3}{c}{Learned} & \multicolumn{3}{c}{Model-based} & \multicolumn{2}{c}{Total Examples}\\
        & F1 & Specificity & Sensitivity & F1 & Specificity & Sensitivity & \%True & \%False \\
        \midrule
        Left of & 0.911 & 97.3\% & 85.6\% & 0.914 & 98.8\% & 85.1\% & 13.9\% & 81.6\% \\
        Right of & 0.929 & 96.9\% & 89.3\% & 0.885 & 99.0\% & 80.0\% & 14.0\% & 86.0\% \\
        In front of & 0.759 & 97.0\% & 62.3\% & 0.660 & 91.9\% & 51.4\% & 4.6\% & 95.4\% \\
        Behind & 0.653 & 97.8\% & 49.0\% & 0.852 & 86.9\% & 83.4\% & 5.0\% & 95.0\% \\
        Above & 0.867 & 99.5\% & 76.7\% & 0.784 & 98.4\% & 65.1\% & 4.3\% & 95.7\% \\
        Below & 0.822 & 99.4\% & 70.0\% & 0.756 & 98.1\% & 61.5\% & 4.3\% & 95.7\% \\
        Near & 0.869 & 93.7\% & 81.0\% & 0.825 & 98.4\% & 71.1\% & 29.0\% & 71.0\% \\
        Touching & 0.923 & 97.9\% & 88.5\% & 0.418 & 99.2\% & 26.5\% & 12.7\% & 96.3\% \\
        Centered & 0.659 & 97.6\% & 49.7\% & 0.035 & 100.0\% & 1.8\% & 5.6\% & 84.4\% \\
        \bottomrule
    \end{tabular}
    \vskip 6pt
    \caption{Accuracy of the predicate predictor $p_\rho$ in held-out randomly-generated simulated test scenes. Some predicates in our scenes can be very difficult due to clutter and occlusions, such as \textit{in front of}.}
    \label{tab:dataset-results-big}
\end{table}

\revised{
Table~\ref{tab:dataset-results-big} shows extra results from our baseline comparison. This includes sensitivity and specificity, as well as prevalence in the dataset overall. We can see here that for the most part our model is better than the baseline, although not always. We should note, though, that even in the case where it is not better, it's still an improvement since we do not need to implement some complex logic to define the predicates -- as mentioned above in Sec.~\ref{sec:directional}, the logic for computing these arrangements isn't trivial.
}

\end{document}